\documentclass{article}

\usepackage[final]{corl_2020} 

\usepackage{url}            
\usepackage{amsfonts}       
\usepackage{nicefrac}       
\usepackage{microtype}      
\usepackage{bm}
\usepackage{tabularx}
\usepackage{enumitem}
\usepackage[olditem,oldenum]{paralist}
\usepackage{booktabs}
\usepackage{makecell}
\usepackage{adjustbox}
\usepackage{textcomp}
\usepackage{flafter}
\usepackage{pifont}
\newcommand{\cmark}{\ding{51}}%
\newcommand{\xmark}{\ding{55}}%

\usepackage{listings}
\usepackage{color}
\usepackage{bm}

\ifx \@etal \@empty  \fi
\ifx \@eg \@empty \newcommand{\eg}{e.g.,~} \fi
\ifx \@ie \@empty \newcommand{\ie}{i.e.,~} \fi
\ifx \@etc \@empty  \fi

\newcommand{\eg}{e.g.,~}
\newcommand{\ie}{i.e.,~}

\newcommand{\figref}[1]{Figure~\ref{#1}}

\newcommand{\secref}[1]{Section~\ref{#1}}

\newcommand{\tabref}[1]{Table~\ref{#1}}

\definecolor{bgCode}{rgb}{0.94, 0.94, 1.0}
\newcolumntype{Y}{>{\centering\arraybackslash}X}

\newcommand{\coda}{Coda}

\newcommand{\xhdr}[1]{\vspace{3pt}\noindent\textbf{#1}}

\title{Sim-to-Real Transfer for\\Vision-and-Language Navigation}

\author{
	Peter Anderson\textsuperscript{1}\thanks{Now at Google.} \hspace{5mm}
	Ayush Shrivastava\textsuperscript{1} \hspace{5mm}
	Joanne Truong\textsuperscript{1} \hspace{5mm}
	Arjun Majumdar\textsuperscript{1} \\
	\textbf{Devi Parikh\textsuperscript{1,2}} \hspace{5mm}
	\textbf{Dhruv Batra\textsuperscript{1,2}} \hspace{5mm}
	\textbf{Stefan Lee\textsuperscript{3}} \vspace{2mm} \\
	$^1$Georgia Institute of Technology ~
	$^2$Facebook AI Research ~ 
	$^3$Oregon State University
}

\begin{document}
\maketitle


\begin{abstract}
   We study the challenging problem of releasing a robot in a previously unseen environment, and having it follow unconstrained natural language navigation instructions. Recent work on the task of Vision-and-Language Navigation (VLN) has achieved significant progress in simulation. To assess the implications of this work for robotics, we transfer a VLN agent trained in simulation to a physical robot. To bridge the gap between the high-level discrete action space learned by the VLN agent, and the robot's low-level continuous action space, we propose a subgoal model to identify nearby waypoints, and use domain randomization to mitigate visual domain differences. For accurate sim and real comparisons in parallel environments, we annotate a 325m\textsuperscript{2} office space with 1.3km of navigation instructions, and create a digitized replica in simulation. We find that sim-to-real transfer to an environment not seen in training is successful if an occupancy map and navigation graph can be collected and annotated in advance (success rate of 46.8\% vs. 55.9\% in sim), but much more challenging in the hardest setting with no prior mapping at all (success rate of 22.5\%).
\end{abstract}

\section{Introduction}

We study the challenging problem of releasing a robot in a previously unseen environment, and having it follow unconstrained natural language navigation instructions. Most previous evaluations of instruction-following robots either focus on smaller table-top environments~\cite{guadarrama2013grounding,misra2016tell,paul2016efficient}, or are evaluated in simulation~\cite{chen2011learning, tellex2011understanding, matuszek2013learning, bisk2016natural,chaplot2017gated,kollar2010toward,macmahon2006walk,mei2016listen,vogel2010learning}. However, performing only component-level evaluation (\eg of the instruction parser) or evaluating only in simulation neglects real-world sensing, actuation and localization errors and the challenges of integrating complex components, which may give a misleading impression of progress. Therefore, leveraging the cumulative advances of previous authors studying Vision-and-Language Navigation (VLN) in simulation~\cite{mattersim,wang2018reinforced,wang2018look,ma2019selfmonitoring,ma2019theregretful,fried2018speaker,backtranslate2019,ke2019tactical}, we transfer a VLN agent trained in simulation on the R2R dataset~\cite{mattersim} to a physical robot and complete one of the first full system evaluations of a robot following unconstrained English language directions in an unseen building.

\begin{figure}[t]
	\begin{center}
		\begin{minipage}[]{0.62\linewidth}
			\begin{center}
				\includegraphics[width=\linewidth]{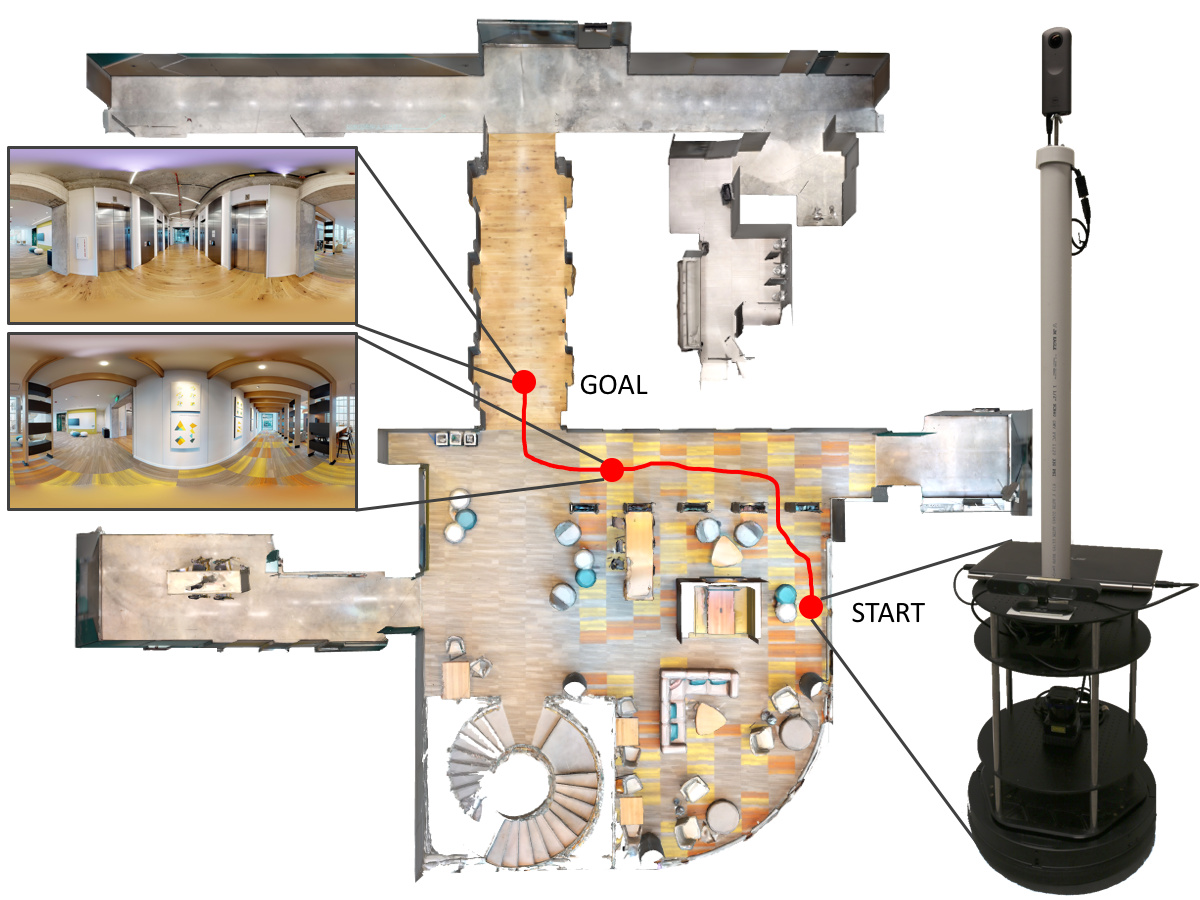}
			\end{center}
		\end{minipage}
		\begin{minipage}[]{0.35\linewidth}
			\begin{center}
				\scriptsize
				\setlength\tabcolsep{3pt}
				\begin{tabularx}{\linewidth}{X}
					\includegraphics[trim=0 0cm 0 0cm, clip,width=1\linewidth]{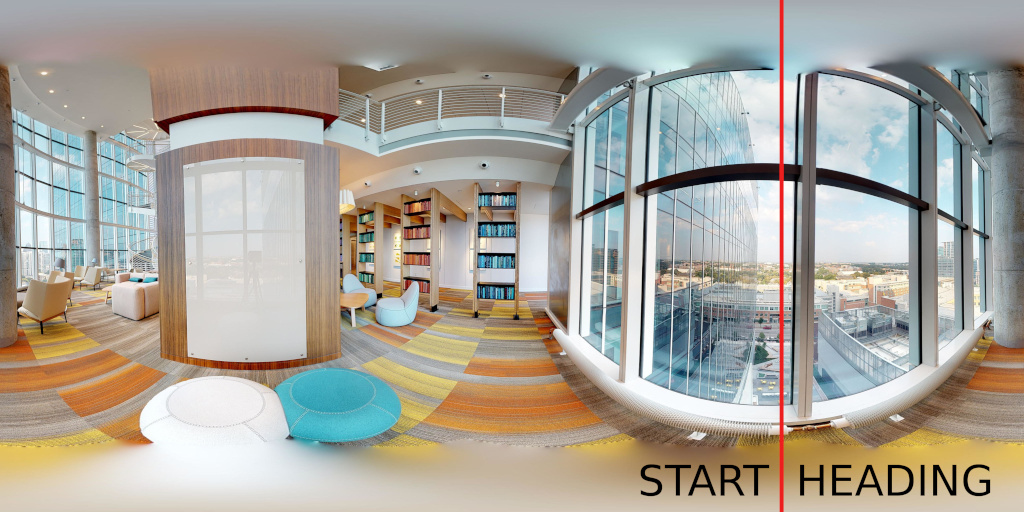} \\
					Go between the first and second bookshelves, turn to your left and walk straight down the hallway, then you should turn to your right at the hallway with the elevators and stop when the fire extinguisher box is on your left. \\
					\midrule
					Walk between the two bookshelves, turn left, walk past the last set of pictures on the wall, turn right and wait by the elevators.\\
					\midrule
					Turn left and head toward and past the blue bookcase. Turn left again and walk down the long hallway until you get to the opening on the right. Turn right and head toward the elevators and you're there.\\
				\end{tabularx}
			\end{center}
		\end{minipage}
	\end{center}
	\caption{We transfer a VLN agent trained in simulation to a physical robot (center) placed in a 325m\textsuperscript{2} office environment not seen in training (left). Our experiments compare instruction-following performance in parallel simulator and physical environments over 111 unconstrained natural language navigation instructions (\eg the 3 instructions on right corresponding to the red target trajectory). }
	\label{fig:concept}
\end{figure}

As illustrated in \figref{fig:concept}, VLN~\cite{mattersim} is a formulation of the instruction-following problem that requires an agent to interpret a natural-language instruction and then execute a sequence of actions to navigate efficiently from the starting point to the goal in a previously unseen environment. In existing working studying VLN in simulation, the agent's action space is typically defined in terms of edge traversals in a navigation graph, where nodes are represented with 360\textdegree{} panoramic images (on average 2.1m apart) and edges indicate navigable paths between these panoramas. In effect, high visual fidelity comes at the cost of low control fidelity. Given this limitation, a major challenge with sim-to-real transfer of a VLN agent is 
bridging the gap between the high-level discrete action space learned by the agent and the low-level continuous physical world in which the robot operates.

To address this challenge, we propose a subgoal model that, conditioning on both 360\textdegree{} RGB images and laser scans, identifies a set of navigable nearby waypoints that can be evaluated by the VLN agent as high-level action candidates. To minimize domain differences between sim and real, the subgoal model is trained on the same navigation graphs (from the R2R / Matterport3D~\cite{Matterport3D} dataset) as the VLN agent. To support navigation to the waypoint selected by the VLN agent, we assemble a classical navigation stack based on Robot Operating System (ROS)~\cite{quigley2009ros}, incorporating a standard Simultaneous Localization and Mapping (SLAM) implementation~\cite{grisetti2007improved} along with obstacle avoidance and path-planning routines. For sim-to-real experiments we use a TurtleBot2 mobile robot equipped with a 2D laser scanner and a 360\textdegree{} RGB camera. To mitigate the impact of visual differences between sim and real, we train the VLN agent and the subgoal model using domain randomization~\cite{tobin2017domain}.

Evaluations are conducted in a 325m\textsuperscript{2} physical office environment we refer to as \coda{}. Since our focus is evaluating whether progress on VLN in simulation can be parlayed into progress in robotics, we digitize and annotate the \coda{} environment to provide a reliable comparison of sim and real performance in parallel unseen test environments. Specifically, a Matterport camera is used to scan and reconstruct \coda{} and the resulting assets are imported into the Matterport3D simulator. We then construct a dataset of English language instruction-trajectory pairs in \coda{} using Amazon Mechanical Turk (AMT) and following the R2R data collection protocols. In total, we collect 111 navigation instructions representing 1,334m of language-guided trajectories. 

Experimentally, we complete two full physical evaluations in \coda{}, testing two settings: `with map' in which an occupancy map and navigation graph are collected and annotated in advance, and `no map' in which the robot performs SLAM from scratch each time a new instruction is received. In all evaluations the language inputs and the visual appearance of the environment are previously unseen. We show that sim-to-real transfer is reasonably successful in the `with map' setting (success rate of 46.8\% vs. 55.9\% in simulation, with reductions in success rate attributed 3.9\% to remaining visual domain differences and 5.2\% to viewpoint differences). However, in the hardest setting with no prior mapping, sim-to-real transfer is much less reliable (success rate of 22.5\%) due to subgoal prediction errors which fail to fully abstract the differences in the agent's action space between sim and real.


\xhdr{Contributions.} In summary, we achieve the first sim-to-real transfer of a VLN agent trained in simulation on the R2R dataset to a robotic platform. Our main contributions include:
\begin{compactitem}[-]
	\item A new annotated VLN simulator environment corresponding to an accessible physical environment, 
	\item A sim-to-real framework interfacing a trained VLN agent with standard ROS components,
	\item A subgoal prediction model to bridge the gap between the discrete action space learned by the VLN agent and the continuous world, and
	\item An empirical study quantifying the sim-to-real gap in terms of errors due to visual domain differences, viewpoint differences and subgoal prediction errors / action space differences.
\end{compactitem}
We provide code\footnote{\href{https://github.com/batra-mlp-lab/vln-sim2real}{https://github.com/batra-mlp-lab/vln-sim2real}} for digitizing and annotating new VLN environments, plus our sim-to-real framework and subgoal model for deploying VLN agents to robots using ROS.

\section{Related Work}

\xhdr{Instruction-Following Robots} Although natural language command of robots in unstructured environments could be considered a grand challenge of robotics, there is surprisingly little previous work in our setting of interest -- in which a physical robot is released in a building it hasn't seen in training and evaluated on it's ability to execute unconstrained natural language navigation instructions. The most similar settings to ours are (noting major differences from ours in parentheses): the voice-commandable wheelchair of \citet{hemachandra2015learning} (relies on artificial landmarks), the quadcopters of \citet{huang2010natural} and \citet{blukis2019learning} (evaluated in the training environment), and the voice-commanded robot teammate of \citet{oh2016integrated} (outdoors). Most other evaluations of language-guided robots have been limited to only a handful of instruction commands \cite{codevilla2018end, Patki2019LanguageguidedSM} or a handful of object types \cite{tucker2020learning,RobotSlang20}, or they focus on manipulation rather than navigation \cite{fahnestock2019language, paul2018temporal}.  

\xhdr{Vision-and-Language Navigation (VLN)} In the VLN task~\cite{mattersim}, an agent is placed in a photo-realistic simulation of an indoor environment and given a natural language navigation instruction describing the path to the goal. To reach it, agents must learn to ground language instructions to both visual observations and actions. In the standard setting, the test environments are unseen during training and no prior exploration is permitted. Despite the task's difficulty, recent work has seen significant improvements~\cite{wang2018reinforced,wang2018look,ma2019selfmonitoring,ma2019theregretful,fried2018speaker,backtranslate2019,ke2019tactical}, including the use of pragmatic speaker models for trajectory re-ranking and data augmentation~\cite{fried2018speaker,backtranslate2019}, as well as progress estimation~\cite{ma2019selfmonitoring} and backtracking~\cite{ma2019theregretful,ke2019tactical}. 

\xhdr{Pano-Simulators and Datasets} A growing number of simulation environments, tasks and datasets have been proposed based on situated panoramic images. For example, building on R2R/Matterport3D~\cite{mattersim,Matterport3D}, annotations for vision-and-dialog navigation~\cite{thomason:corl19}, asking for help~\cite{nguyen2019hanna}, remote embodied referring expressions~\cite{reverie}, and multilingual VLN~\cite{rxr,yan2019crosslingual} have been released. In the outdoor setting, several panoramic image datasets have been proposed including StreetLearn~\cite{mirowski2018learning,mehta2020retouchdown} and SEVN~\cite{weiss2019navigation}, giving rise to language navigation datasets such as TouchDown~\cite{Chen19:touchdown}, Talk2Nav~\cite{vasudevan2019talk2nav} and RUN~\cite{pazargaman2019run}. With the increasing interest in training embodied agents in panoramic image environments, there is an urgent need to investigate the transfer of these agents to real physical platforms.

\section{Sim-to-Real Experimental Setting}

\xhdr{\coda{} Test Environment}\label{sec:coda}
For evaluation of the VLN robot we select \coda{} as the unseen test environment. \coda{} is a 325m\textsuperscript{2} collaborative shared space in a commercial office building. As a shared space, \coda{} is devoid of personal items such as papers, posters, photos, bags and computing devices. While this eliminates some interesting visual clutter, it also helps minimize the drift between the static simulator and the real physical environment over time. Visual diversity is enhanced by the variety of rooms included, such as an elevator lobby, several long corridors, a lounge area and various break-out spaces (refer floorplan in \figref{fig:concept}), while floor-to-ceiling glass walls and windows provide reflections and changing lighting that makes the space particularly challenging for robotic vision.

\xhdr{Simulator Construction} To accurately establish the sim-to-real gap in the unseen test environment, we construct a parallel simulator environment by reconstructing \coda{} with a Matterport3D Pro 2 camera and the Matterport3D web services. We download the resulting pointcloud and textured mesh, plus the equirectangular panoramic image and pose at each of the camera 65 viewpoints, and a `visibility graph' indicating which pairs of camera viewpoints are mutually visible. After excluding 6 viewpoints located on stairs or above furniture (which are unreachable by our robot), following \citet{mattersim} we construct a navigation graph from the visibility graph by excluding edges with length greater than 5m. The resulting navigation graph, panoramic images and viewpoint poses are then imported into the Matterport3D simulator to create the \coda{} simulator environment.

\xhdr{Navigation Instructions} To collect navigation instructions for \coda{}, we follow \citet{mattersim} by sampling trajectories that are the shortest path between two points and then asking annotators to describe these paths using an immersive 3D web interface. To validate instruction quality, we then ask a different annotator to follow each instruction using a similar interface. In total we sample 37 trajectories and collect four English language navigation instructions for each trajectory using Amazon Mechanical Turk (AMT). After discarding the instruction with the highest follower navigation error for each trajectory, the final \coda{} dataset contains 111 instructions, representing a total of 1,334m of language-guided navigation with an average human follower success rate of 93\% (vs. 86\% for the R2R test set). As indicated by the example in \figref{fig:concept}, the resulting trajectories and instructions are similar in length and style to R2R, with an average of 25 words per instruction (vs. 29 for R2R). The vocabulary size of the \coda{} instructions is in the 44th percentile of R2R environment vocabularies (based on repeatedly randomly sampling 37 paths / 111 instructions per R2R environment, and skipping R2R environments with insufficient instructions). This suggests that the language is as diverse as many of the environments in the original dataset. Qualitatively, the instructions mention a variety of objects (\eg water fountain, art, sectional sofa), along with attributes such as color (12 mentioned), state (\eg open, closed, hanging), size (\eg small, big), and composition (\eg wood, glass, cement) using both allocentric and egocentric reference frames.

\xhdr{Robot Platform} We conduct experiments using a low-cost TurtleBot2 robot consisting of a Kobuki mobile base, an Asus Xion Pro Live RGBD camera, and an Asus netbook. Since virtually all VLN agents have been developed using 360\textdegree{} vision, we additionally equip the robot with a Ricoh Theta V 360\textdegree{} consumer-grade RGB camera. For the most direct sim-to-real comparison we mount the Theta V camera 1.35m above ground level -- the same height we set the Matterport camera tripod when scanning and reconstructing \coda{}. Lastly, for obstacle avoidance and mapping we mount a 270\textdegree{} Hokuyo 2D laser scanner with 30m range at 0.24m above ground level. The robot runs ROS-kinetic~\cite{quigley2009ros}. We use standard ROS / TurtleBot packages such as gmapping, amcl and move\_base as well as PyTorch~\cite{paszke2017automatic}. During evaluation all code executes on the robot, except PyTorch ROS nodes (the VLN agent and our subgoal model) which are run remotely.

\xhdr{Evaluation Metrics} We use standard VLN metrics for evaluation in both sim and real, with the aim of testing whether performance on the robot can match the results in simulation for the same unseen test environment. In all experiments the agent must terminate the episode as near to the goal position as possible. An episode is considered successful if the navigation error (defined as the distance between the agent's final position and the goal position) is less than 3m. We report average values for trajectory length (\texttt{TL}), navigation error (\texttt{NE}), success rate at reaching the goal (\texttt{SR}), oracle success rate (\texttt{OS}) -- defined as the agent's success rate at the closest point on the agent's trajectory to the goal, Success weighted by (normalized inverse) Path Length (\texttt{SPL})~\cite{evaluation2018} and both Normalized and Success-weighted Dynamic Time Warping (\texttt{NDTW} and \texttt{SDTW})~\cite{magalhaes2019effective}. \texttt{SPL} and \texttt{SDTW} are summary measures of navigation performance that balance success against trajectory efficiency and fidelity (\ie similarity to the ground-truth path), while \texttt{NDTW} measures path fidelity irrespective of success (higher is better for each). When calculating \texttt{NDTW} and \texttt{SDTW} we represent both sim and robot trajectories using 100 equally spaced points to eliminate sampling differences.

\xhdr{Robot Pose Tracking}
To compute these evaluation metrics we must know the agent's pose at all points in time. In the simulation this is available. To track the robot's pose, we first teleop the robot through \coda{} to construct a 0.05m resolution occupancy map using the laser scanner and the ROS gmapping SLAM package. We then register this map to the Matterport3D coordinate frame. For evaluation purposes we use the particle filter provided by the ROS amcl package to track the robot's pose during experiments. We estimate that the error in this pose estimate is typically an order of magnitude less than the 3m radius used for determining success.

\begin{figure*}[t]
	\begin{center}
		\adjustbox{trim={0} {0} {0} {0},clip}%
		{\includegraphics[width=1.0\linewidth]{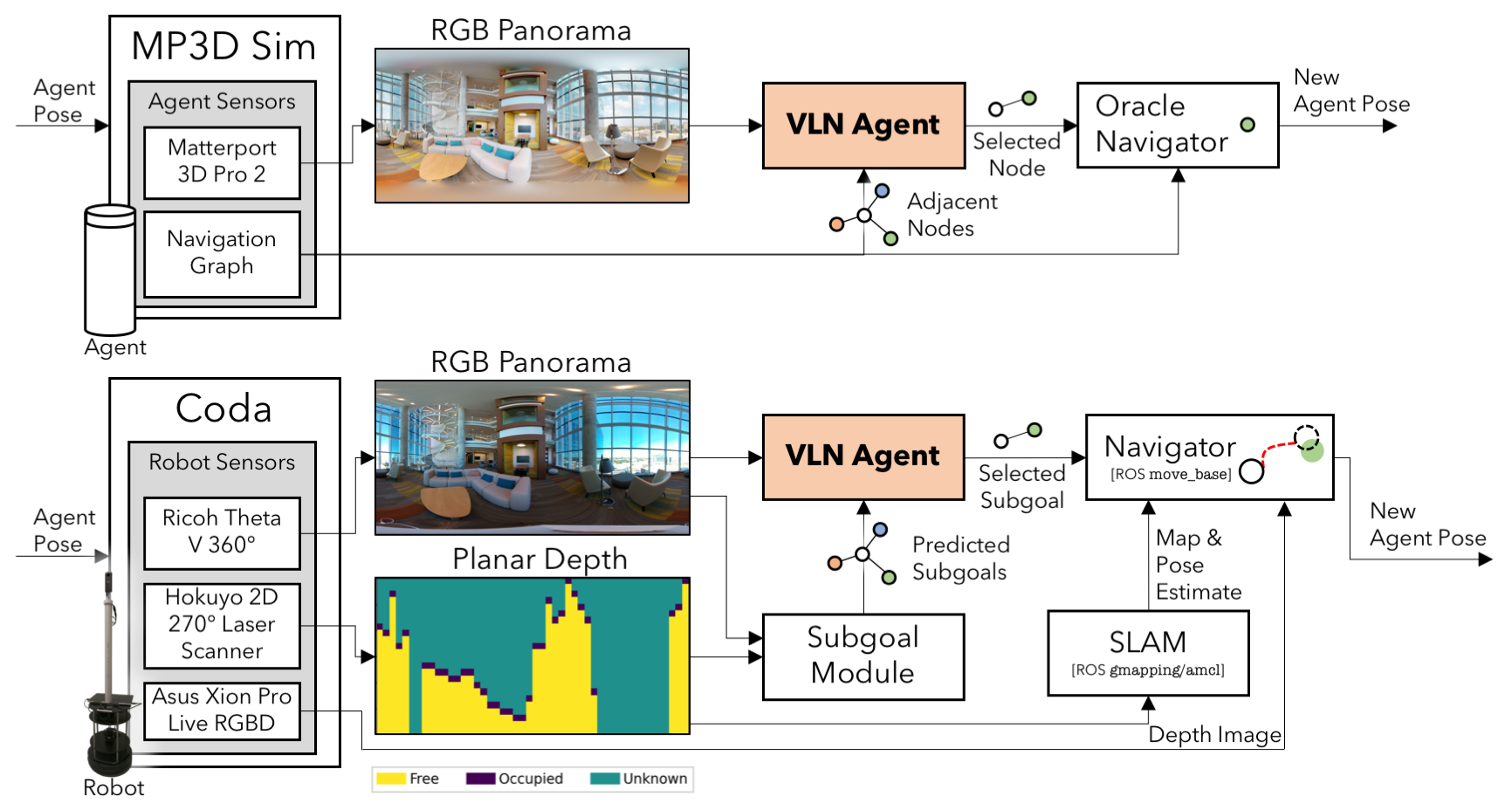}}
	\end{center}
	\caption{We evaluate a VLN agent trained in simulation in matching previously unseen sim (top) and real (bottom) environments. Our sim-to-real framework includes domain randomization, a subgoal model to bridge the action space differences between sim and real (removing the reliance on the simulator navigation graph), and standard ROS components for SLAM and navigation. }
	\label{fig:system}
\end{figure*}

\section{Adapting a VLN Agent to a Physical Robot}

\xhdr{VLN Agent} We now detail our approach to sim-to-real transfer of a VLN agent onto the robot.
We start with an existing VLN agent~\cite{backtranslate2019} that achieved state-of-the-art performance on the unseen environments in the R2R test split. Typical of most recent work on this task, this agent is trained using a (simulated) panoramic 360\textdegree{} RGB sensor, processing the entire visual context at each step, and a highly-abstracted discrete action space provided by the Matterport3D simulator. Actions are defined by neighboring panoramic image viewpoints, which are on average 2.1m away. The agent processes the visual representations in the direction of those locations, and then once an action has been selected the simulator teleports the agent to the new viewpoint (refer \figref{fig:system}). To transfer this agent to a robot we must address differences in the visual domain and action space, plus navigation and localization errors that are not experienced in training. We discuss these challenges in turn.

\xhdr{Visual Domain Adaptation} As illustrated by the RGB panorama differences in \figref{fig:system}, the robot's Ricoh Theta V camera is consumer-grade with limited dynamic range compared to the Matterport camera, resulting in a loss of visual detail and a 6.4\% reduction in success rate due to visual domain differences (refer \secref{sec:results}). 
To address this problem, domain adaptation algorithms typically require a large set of target domain images -- in this case Theta V panoramic images captured in a variety of indoor environments -- which are non-trivial to collect. In preliminary experiments, simple alternatives such as histogram matching to align the Theta V colorspace with the Matterport camera were not effective. We therefore selected an approach akin to domain randomization \cite{tobin2017domain}, in which we applied random color jitter to each panorama when training the VLN agent in the Matterport simulator. Brightness, contrast and saturation were varied by a factor of 0.3 and hue was varied by a factor of 0.01, with these parameters determined through visual inspection.

\begin{figure}[t]
	\begin{center}
		\small
		\setlength\tabcolsep{10pt}
		\begin{tabularx}{\linewidth}{XX}
			\adjustbox{trim={.04\width} {.05\height} {.02\width} {.1\height},clip}%
			{\includegraphics[width=0.56\columnwidth]{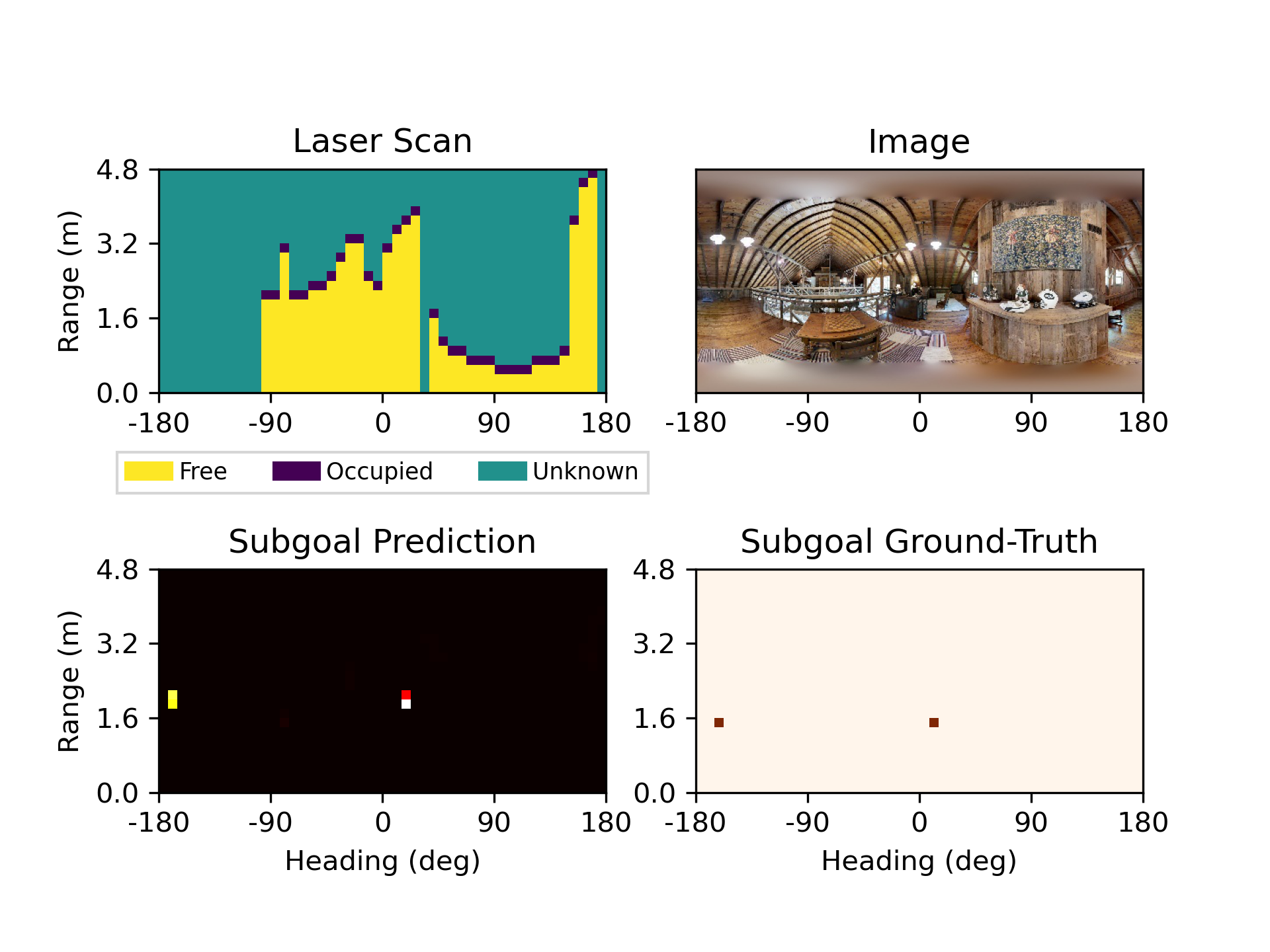}} &
			\includegraphics[trim=1.4cm 0cm 2cm 0, clip,width=0.44\columnwidth]{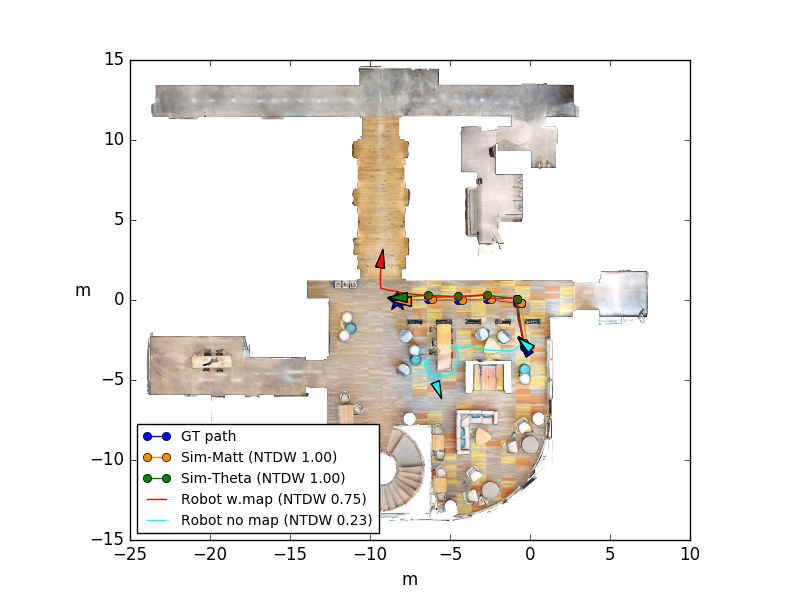}
		\end{tabularx}
	\end{center}
	\caption{Left: To predict the next set of potential waypoints, or subgoals, we combine a radial occupancy map representation of the laser scan (top-left) with pretrained CNN features from the panoramic image (top-middle) in a U-Net~\cite{ronneberger2015u} architecture. Even with missing range data (green regions) due to the 270\textdegree{} scan, the model generally predicts plausible subgoals learned from the Matterport simulator navigation graphs (bottom-left vs. bottom-middle). However, when the subgoal model fails to predict a valid waypoint (right, where the subgoal model does not predict a waypoint passing between narrow bookshelves) then the robot navigation fails. }
	\label{fig:action-pred}
\end{figure}

\xhdr{Action Space Adaptation} To accommodate the transfer from the highly-abstracted action space of the simulator, which is based on a navigation graph, to the primitive motor control actions of a physical robot, we propose to use a \textit{subgoal model} in conjunction with standard ROS navigation modules. After conditioning on available sensor observations, the subgoal model predicts a set of nearby waypoints, or subgoals, for the VLN agent to choose from. In effect, we provide the VLN agent with an implementation of the simulator's action space and depend on ROS to execute those actions. Our approach is therefore a modular fusion of both classical and learning-based methods.

As illustrated in the example in \figref{fig:action-pred} left, as input to the subgoal model we use the most recent laser scan, to provide an approximate indication of free space, and visual inputs to fill in the gaps and add additional context. We represent the laser scan as a radial occupancy map over range and heading bins, and the visual input using pretrained ResNet-152~\cite{he2016deep} CNN features captured at 12 different headings and 3 different camera evaluations (the same visual feature representation consumed by the VLN agent). The subgoal model is based on a 4-stage U-Net~\cite{ronneberger2015u} architecture that takes the radial occupancy map as input and fuses the visual features in the 3rd downsampling stage. The output of the model is the probability that each laser scan heading and range bin contains a waypoint.

The subgoal model is trained and validated on viewpoints from the Matterport3D dataset \cite{Matterport3D}. Since the dataset does not include planar laser scans, we simulate the output of a 270\textdegree{} laser scan at each viewpoint by measuring the range to the reconstructed matterport mesh. The model is trained to minimize the sinkhorn divergence \cite{cuturi2013sinkhorn} between predicted waypoint locations and the location of neighboring viewpoints in the navigation graph, after removing neighboring viewpoints that would require traversing stairs. Sinkhorn divergence is a smoothed approximation to earth mover's distance that we find to be more effective than cross-entropy, which does not respect the underlying metric space. At test time in the \coda{} environment we use a confidence threshold to select the final set of waypoints (up to a maximum of 5) that are provided to the VLN agent. Using the Theta V camera and evaluating at each viewpoint in the Coda navigation graph we find that 29\% of predicted waypoints are within 0.5m of a ground-truth neighboring viewpoint (60\% within 1m, 74\% within 1.5m).


\xhdr{Obstacle Avoidance and Path Planning}
For obstacle detection we rely on the depth camera and the 270\textdegree{} laser scanner. These sensors are complimentary since, although the camera only has a narrow horizontal field of view, it senses overhanging obstacles such as table tops that may be missed in the narrow planar sweep of the laser scanner (and could be impacted by our tall robot). We rely on the ROS move\_base navigation module to integrate these observations, maintain local and global costs maps, and to plan and execute motion to the waypoint selected by the VLN model.

\xhdr{Verification} 
Porting an existing VLN agent to ROS required significant refactoring of the original codebase. To verify our implementation we created mock ROS nodes for the robot's panoramic camera, the subgoal model, and the move\_base navigation package using images from the simulator and the simulator navigation graph. This allowed us to run our robot code on simulator data and verified that the performance matched the original R2R-EnvDrop model~\cite{backtranslate2019} within 1\%. Given the large number of different VLN agents and architectures that have been proposed~\cite{wang2018reinforced,wang2018look,ma2019selfmonitoring,ma2019theregretful,fried2018speaker,backtranslate2019,ke2019tactical}, we hope that our framework will ease the sim-to-real burden in future work. As shown in \figref{fig:system}, this codebase acts as a ROS-based harness around the standard VLN agent api adopted by the community.

\section{Results and Analysis}
\label{sec:results}

In this section, we report sim-to-real results and analysis for the VLN robot. We first characterize the difficulty of the \coda{} environment relative to R2R, and examine the importance of visual domain adaptation. We then consider two sim-to-real settings: the first in which the robot is provided with a laser-scanned occupancy map of the environment and the simulator navigation graph (`with map'), and the second in which the environment is completely unseen at the beginning of each episode, \ie there is no provided map or navigation graph and the robot's SLAM map is reset each time an instruction is received (`no map'). No aspect of our system is trained or validated in the \coda{} environment (in simulation or in reality) in either setting. All experiments are conducted in daylight and furniture is restored to its original position before each evaluation. This does not assist the robot (which, in our hardest setting, experiences the environment as previously unseen each time an instruction is received). Rather, this minimizes drift between the simulator and the physical environment, enabling us to more accurately characterize sim-to-real performance.

\xhdr{How challenging is \coda{}?}
We first establish the relative difficulty of the \coda{} environment compared to existing environments in the Matterport3D~\cite{Matterport3D} / R2R~\cite{mattersim} dataset. We report the performance of the R2R-EnvDrop~\cite{backtranslate2019} VLN agent in the \coda{} simulator compared to R2R val and test. As illustrated in \tabref{tab:hard}, after training on the R2R training set we achieved a 49.9\% success rate on the R2R val-unseen set, and a 49.2\% success rate on R2R test. On \coda{} (sim), the agent's success rate is slightly higher at 55.9\%. This suggests that, although the \coda{} dataset was collected by a different camera operator, and a different pool of AMT workers at a different time to R2R, the collection protocol has been faithfully followed and the dataset is `in-domain' with respect to R2R. Accordingly, we treat \coda{} sim performance (row 4) as a baseline to investigate sim-to-real transfer.

\begin{table}[h]
	\small
	\begin{center}
		\begin{tabularx}{\columnwidth}{llYYYYYYY}
			\midrule
			& Split & \textbf{\texttt{TL}} (m) & \textbf{\texttt{NE}} (m) & \textbf{\texttt{OS}} (\%) & \textbf{\texttt{SR}} (\%) & \textbf{\texttt{SPL}} &  \textbf{\texttt{SDTW}}  & \textbf{\texttt{NDTW}} \\
			\midrule
			\small\texttt{1} & R2R Val-Seen           & 9.70  & 4.56 & 62.0 & 57.5 & 55.0 & 49.6 & 60.8 \\
			\small\texttt{2} & R2R Val-Unseen         & 9.18  & 5.42 & 55.4 & 49.9 & 47.0 & 44.5 & 55.2 \\
			\small\texttt{3} & R2R Test               & 9.52  & 5.57 & 54.9 & 49.2 & 46.8 & - & - \\
			\small\texttt{4} & \coda{}                & 11.22 & 4.98 & 59.5 & 55.9 & 53.6 & 44.0 & 54.9 \\
			\midrule
		\end{tabularx}
	\end{center}
	\caption{Characterizing the difficulty of the \coda{} simulator environment using the EnvDrop agent~\cite{backtranslate2019}. Performance in \coda{} is slightly higher than the R2R test set on average. R2R Val-Unseen, R2R Test and \coda{} are previously unseen visual environments.}
	\label{tab:hard}
\end{table}

\xhdr{How important is visual domain adaptation?}
To address this question, we replaced the Matterport panoramic images in the \coda{} simulator using images captured with the robot's Ricoh Theta V camera from the same viewpoint locations. This allows us to isolate the importance of visual domain adaptation in the absence of other factors. As foreshadowed in \figref{fig:system}, switching to the cheaper camera (row 5 vs. row 4 in \tabref{tab:results}) causes a 6.4\% drop in success rate and a 6.2\% drop in SPL, which is reduced to a 3.9\% and 4.3\% drop respectively (row 6 vs. row 4) when we retrained the agent using visual domain adaptation. To make these results as reliable as possible, rows 5 and 6 are averaged over 3 sets of Theta V images that were captured on different days, exhibiting a variation in success rate of $\pm$1.9\% without domain adaptation and $\pm$1.4\% with domain adaptation.

\begin{table}[h]
	\setlength{\tabcolsep}{.3em}
	\small
	\begin{center}
		\begin{tabularx}{\columnwidth}{llYYYYYYYYYY}
			\midrule
			& Setting & Camera & Adapted & Map & \textbf{\texttt{TL}} (m) & \textbf{\texttt{NE}} (m) & \textbf{\texttt{OS}} (\%) & \textbf{\texttt{SR}} (\%) & \textbf{\texttt{SPL}}  &  \textbf{\texttt{SDTW}}  & \textbf{\texttt{NDTW}} \\
			\midrule
			\small\texttt{4} & Sim    &  Matterport   & - & - & 11.22 & 4.98 & 59.5 & 55.9 & 53.6  & 44.0 & 54.9 \\
			\small\texttt{5} & Sim    &  Theta V   & \xmark & - & 11.21 & 5.71 & 55.3 & 49.5 & 47.4 & 39.5 & 52.0 \\
			\small\texttt{6} & Sim    &  Theta V   & \cmark & - & 11.38 & 5.74 & 59.2 & 52.0 & 49.3 & 42.6 & 54.0 \\
			\midrule
			\small\texttt{7} & Robot  &  Theta V   & \cmark & \cmark & 11.32     & 6.04    & 51.4    & 46.8    & 43.9 & 28.7 & 38.5 \\
			\small\texttt{8} & Robot  &  Theta V   & \cmark & \xmark & 8.02 & 6.56 & 26.1 & 22.5 & 21.9 & 13.8 & 30.0 \\
			\midrule
		\end{tabularx}
	\end{center}
	\caption{Sim-to-real performance comparison for our VLN agent in \coda{} over 1,334m of language-guided trajectories. Success rate at reaching the goal (\textbf{\texttt{SR}}) and path fidelity (\textbf{\texttt{NDTW}}) remains relatively high in the robot `with map' setting (row 7), but is reduced in the `no map' setting (row 8). }
	\label{tab:results}
\end{table}

\xhdr{How large is the sim-to-real gap with a map?}\label{sec:map}
In the `with map' setting, we conduct a full evaluation on the robot while also providing a pre-captured laser scan to assist with obstacle avoidance and path-planning, and the simulator navigation graph to provide waypoint candidates. The subgoal model is not used. We consider this setting for two reasons. First, for a robot operating in a single environment it may be reasonable to provide some navigation annotations and to expect the robot to maintain an occupancy map. Second, this setting tests the implicit assumption in graph-based simulators that existing robotics systems are capable of navigating between viewpoints in the graph.

As reported in \tabref{tab:results}, in the `with map' setting the robot achieves an instruction-following success rate of 46.8\% in physical \coda{}, a reduction of 5.2\% compared to using the robot camera in simulation (row 7 vs row 6). During this evaluation over 111 instructions and 1.3km of trajectories we did not experience any collisions or navigation failures, which we define as the robot failing to navigate an edge between panoramic viewpoints in the navigation graph. For context, the environment contains five `squeeze points' with 80-90cm gaps between furniture, compared to which the width of the robot with laptop is around 40cm. We attribute the reduction in performance to a degradation of visual inputs from viewpoints that are slightly out-of-position compared to the simulator. We also note that, despite our best efforts, there were some people present in \coda{} at times during testing, which may also degrade robot performance due to domain differences (images of people are extremely rare in the Matterport3D dataset). Overall we conclude that, at least based on analysis performed in the \coda{} environment, if an occupancy map and navigation graph are collected and annotated in advance, transferring a VLN agent to an inexpensive robot using a classical navigation stack is feasible with around a 9.1\% reduction in success rate (row 7 vs. row 4).

\xhdr{How large is the sim-to-real gap overall?}
\label{sec:nomap}
In the final experimental setting, we revoke the robot's access to the \coda{} occupancy map and the simulator navigation graph. Removing the navigation graph requires the robot to rely on waypoint predictions from the subgoal model for the first time. Removing the occupancy map requires the robot to perform simultaneous localization and mapping (SLAM) from scratch in each episode (every time a new instruction is received), which makes obstacle avoidance and path planning more challenging.\footnote{Note that for pose tracking, we run a separate, isolated ROS navigation stack with map access.} Overall, this setting is indicative of `cold-start' performance in a new environment (which is also the standard VLN evaluation setting in simulation).

As reported in row 8 of \tabref{tab:results}, in the `no map' setting the instruction-following success rate drops a further 24.3\% to 22.5\%. This experiment also exhibits the lowest trajectory similarity with the results of the Matterport3D simulator (\tabref{tab:ndtw}). Without the occupancy map, the robot collided with objects (a table and a chair) for the first time. However, we attribute most of the drop in success rate to differences between the predictions of the subgoal model and the locations of viewpoints in the simulator navigation graph. We found that both false positive and false negative waypoint predictions were evident, e.g., when the robot attempted to navigate down the spiral staircase (false positive), or failed to consider navigating through the narrow `squeeze points' between bookshelves and other furniture (false negatives). Based on these results, we consider VLN sim-to-real transfer in this setting to be an open research problem.

\setlength{\tabcolsep}{.15em}
\begin{table}[h]
	\small
	\begin{center}
		\begin{tabularx}{\columnwidth}{llYYYYYYYY}
			\midrule
			& Platform & Camera & Adaptation & Map & \small\texttt{4} & \small\texttt{5} & \small\texttt{6} & \small\texttt{7} & \small\texttt{8} \\
			\midrule
			\small\texttt{4} & Sim    &  Matterport& -      & -      & 1.00 & 0.78 & 0.66 & 0.41 & 0.33  \\
			\small\texttt{5} & Sim    &  Theta V   & \xmark & -      & 0.78 & 1.00 & 0.70 & 0.42 & 0.33  \\
			\small\texttt{6} & Sim    &  Theta V   & \cmark & -      & 0.66 & 0.70 & 1.00 & 0.44 & 0.33  \\
			\small\texttt{7} & Robot  &  Theta V   & \cmark & \cmark & 0.41 & 0.42 & 0.44 & 1.00 & 0.57  \\
			\small\texttt{8} & Robot  &  Theta V   & \cmark & \xmark & 0.31 & 0.33 & 0.33 & 0.57 & 1.00  \\
			\midrule
		\end{tabularx}
	\end{center}
	\caption{Trajectory similarity between \coda{} experimental settings based on Normalized Dynamic Time Warping (\textbf{\texttt{NDTW}}) \cite{magalhaes2019effective}. \figref{fig:action-pred} right provides an example to contextualize NDTW values. }
	\label{tab:ndtw}
\end{table}


\section{Conclusion and Future Directions}

We attempt the first sim-to-real transfer of a Vision-and-Language Navigation (VLN) agent trained on the R2R dataset to a low-cost robot with 360\textdegree{} vision, using a learned subgoal model and classical SLAM and path-planning routines. We show that, if an occupancy map and navigation graph can be collected and annotated in advance, sim-to-real transfer is largely successful albeit with a $\sim$9\% reduction in instruction following success due to visual domain differences ($\sim$4\%) and viewpoint differences ($\sim$5\%). This is a promising result, suggesting that the language groundings learned by the agent in simulation can transfer to a physical environment not seen in training. However, in the hardest `cold start' setting with no prior mapping of the environment, sim-to-real transfer is much less reliable due to the failure of the subgoal model to predict the same neighboring waypoints in the simulator navigation graph. To narrow the sim-to-real gap in future work, the subgoal model will need to be improved or eliminated, perhaps by training the VLN agent using a low-level action space (e.g., predicting the heading and distance to move). Since the graph-based Matterport3D simulator cannot support these low-level actions, this would require off-policy reinforcement learning algorithms that can learn from a fixed batch of data that has already been gathered \cite{fujimoto2019off}, or alternatively, switching to a simulator that supports continuous motion \cite{xiazamirhe2018gibsonenv, habitat19iccv}, as in recent work from \citet{krantz2020navgraph}. Since these simulators introduce visual artifacts not present in the graph-based simulator, this may exacerbate visual domain differences, although recent work \cite{kadian2019we} demonstrates a promisingly small sim-to-real gap in the context of PointGoal navigation \cite{anderson2018evaluation}.

\clearpage
\acknowledgments{The Georgia Tech effort was supported in part by NSF, AFRL, DARPA, ONR YIPs, ARO PECASE, Amazon. The views and conclusions contained herein are those of the authors and should not be interpreted as necessarily representing the official policies or endorsements, either expressed or implied, of the U.S. Government, or any sponsor. The license for the Matterport3D dataset is available at \href{http://kaldir.vc.in.tum.de/matterport/MP_TOS.pdf}{http://kaldir.vc.in.tum.de/matterport/MP\_TOS.pdf}.}

\bibliography{paper}


\begin{table}
	\begin{center}
		\setlength\tabcolsep{5.0pt}
		\begin{tabularx}{\columnwidth}{lYlYYYlYYY}
			\toprule
			&       &  & \multicolumn{7}{c}{Matterport3D / R2R Dataset}                          \\
			\cmidrule{4-10}
			&       &  & \multicolumn{3}{c}{Train ($n=61$)} &  & \multicolumn{3}{c}{Val-Unseen ($n=11$)} \\
			\cmidrule{4-6} \cmidrule{8-10}
			& \coda{} &  & Min     & Avg    & Max    &  & Min    & Avg    & Max    \\
			\midrule
			Num Viewpoints       &   59    & &  8 &  125       &  345      &        &  20      &  87      &    215    \\
			Navigation Graph Degree     &  3.6 &    & 2.2  &  4.0       &  5.4        &  &    3.1    &   3.8     &   4.9     \\
			Avg Edge Distance (m)    & 2.8  &    &  1.3 &  2.2       &    3.1    &       &  1.8      &   2.2     &    2.8    \\
			Num Instructions      &   111    &  &   6      &  230      &  300      &  &   18     &    214    &   300     \\
			Avg Instruction Length (words)     &    25   &  &   20      &   29     &  35      &  &    22    &   28     &   32     \\
			Avg Trajectory Length (m) &   12.0    &  &    5.3     &   9.7     &    15.0    &  &   6.1     &  9.2      &   11.1     \\
			Avg Trajectory Edges        &  4.8     &  &   3.0      &   4.9     &    5.4    &  &   3.9     &   4.7     &    5.2    \\
			\midrule
		\end{tabularx}
	\end{center}
	\caption{Comparison of per-environment average statistics between \coda{} and R2R, suggesting that \coda{} is fairly typical of environments found in the Matterport3D / R2R dataset.}
	\label{tab:comp}
\end{table}

\begin{figure}[b]
	\begin{center}
		\footnotesize
		\setlength\tabcolsep{3pt}
		\begin{tabularx}{\linewidth}{XX}
			\includegraphics[trim=0 0.5cm 0 0.5cm, clip,width=1\linewidth]{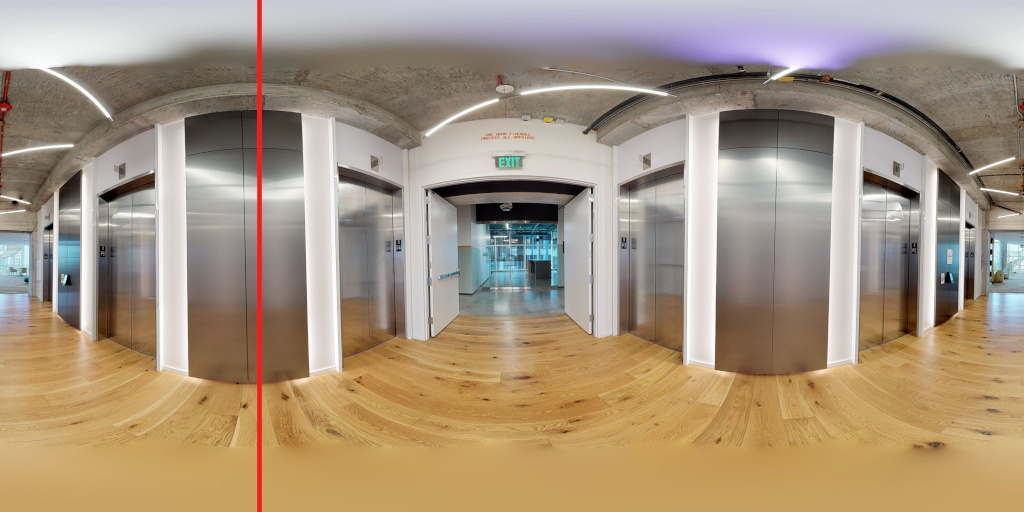} &
			\includegraphics[trim=0 0.5cm 0 0.5cm, clip,width=1\linewidth]{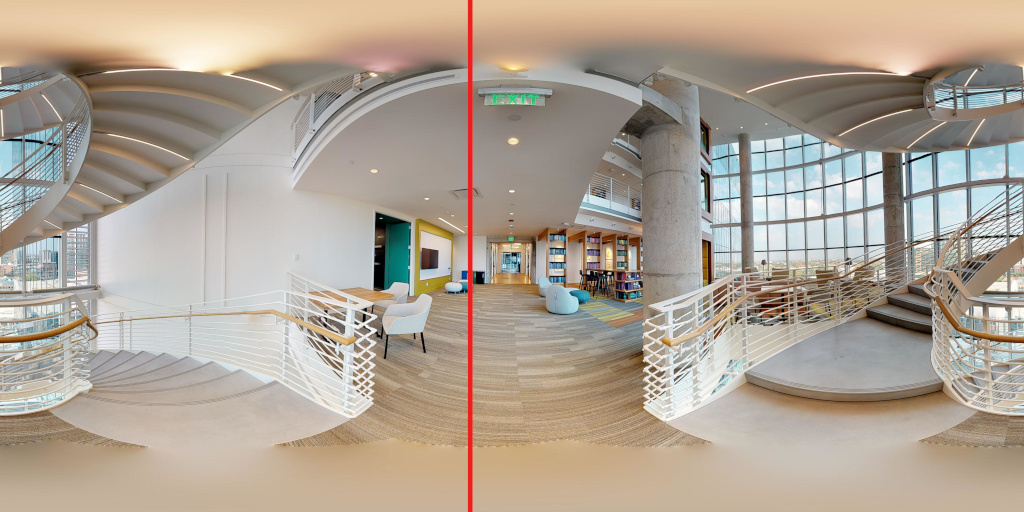} \\
			Turn right, move through the open open double doors. Turn right, move to the end of the hall. Turn right, wait in front of the drinking fountain. 	&
			Walk straight towards the exit, but stop and make a right when you see the wall with art and stop. \\
			\includegraphics[trim=0 0.5cm 0 0.5cm, clip,width=1\linewidth]{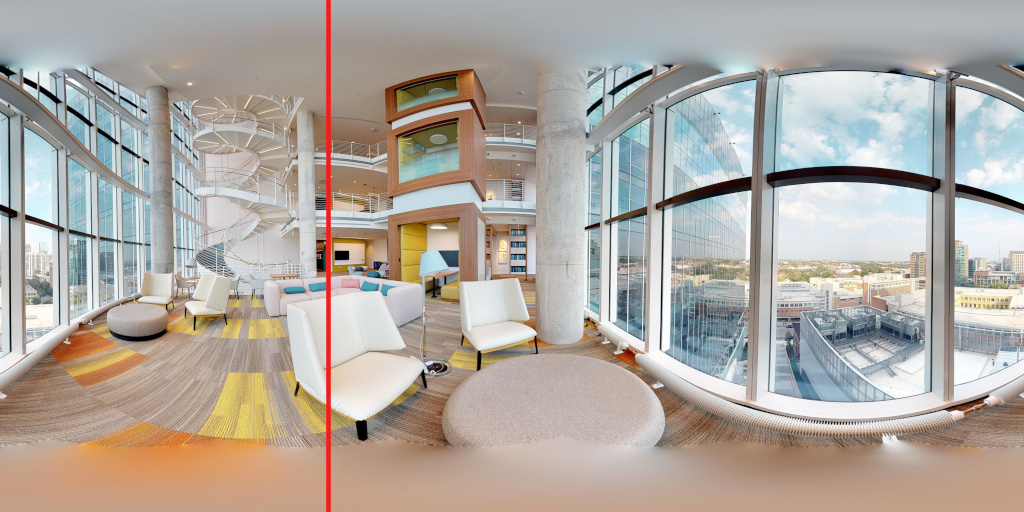} &
			\includegraphics[trim=0 0.5cm 0 0.5cm, clip,width=1\linewidth]{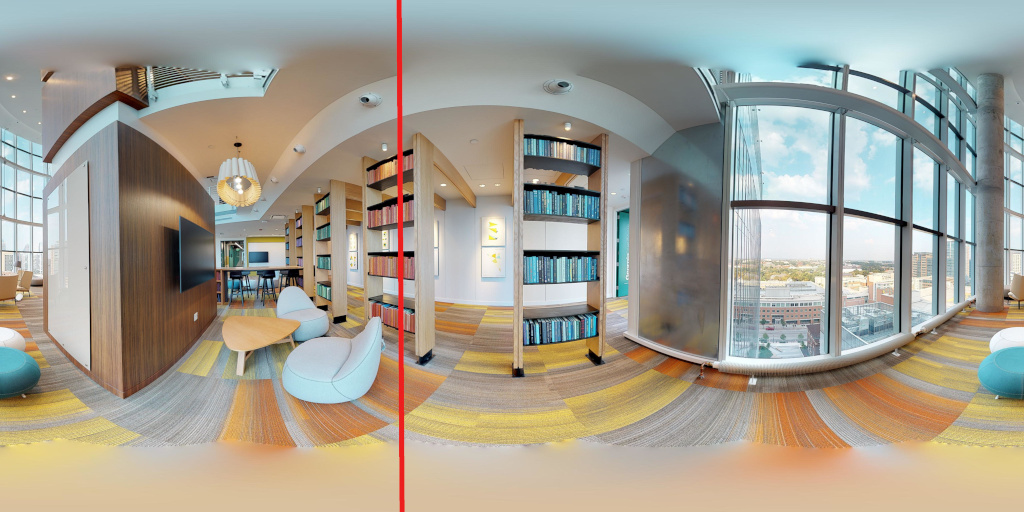} \\
			Make a right in front of the couch, veer straight, then right to proceed past the windows and stop between the first two bookshelves in front of the first hanging piece of art. &
			Go between the first set of bookshelves. Turn left and go straight until you are at the end of the hallway.\\
			\includegraphics[trim=0 0.5cm 0 0.5cm, clip,width=1\linewidth]{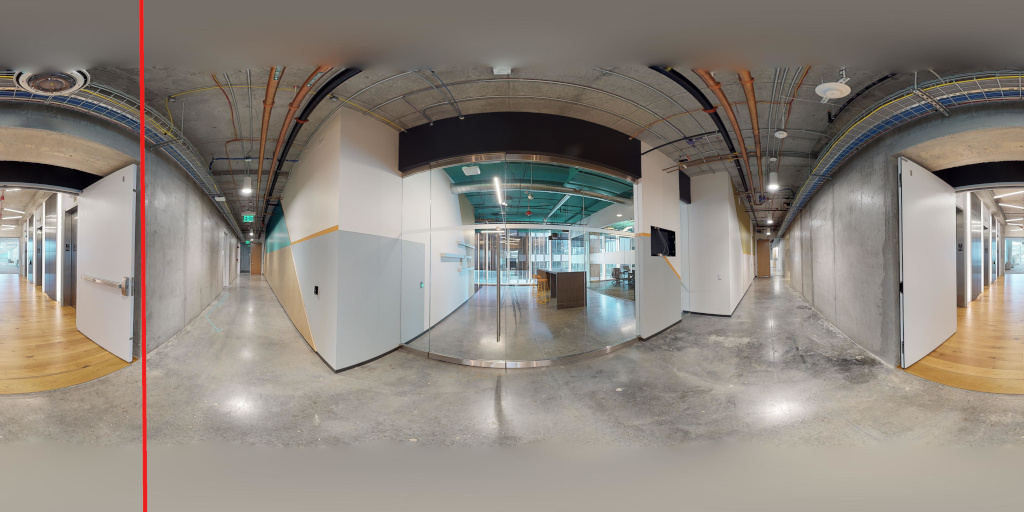} &
			\includegraphics[trim=0 0.5cm 0 0.5cm, clip,width=1\linewidth]{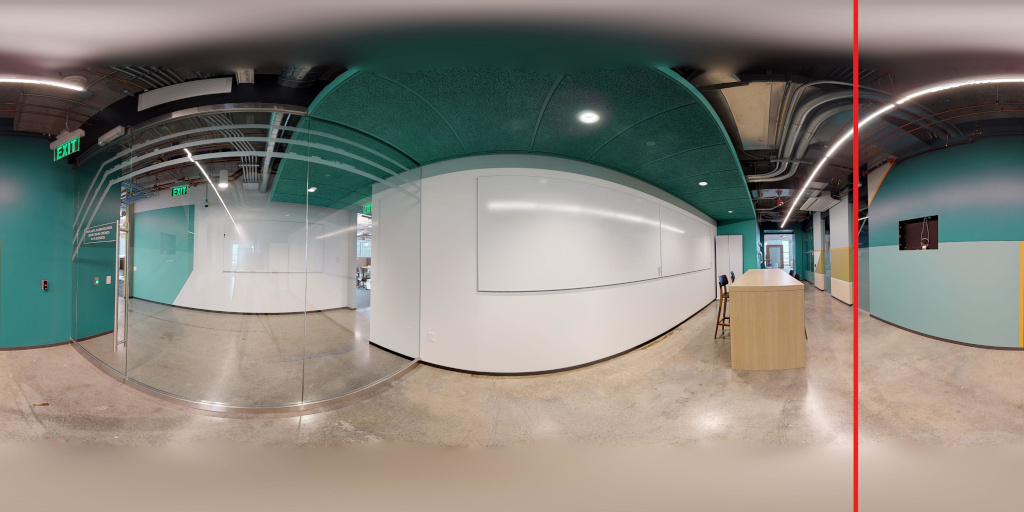} \\
			Turn around and walk to towards the end of the hall. At the intersection, turn and head into the men's room. &
			Continue forward with the whiteboards on your left. Keep walking and you will be in a new room. Stop before you reach the peach-colored couch on your right. \\
		\end{tabularx}
	\end{center}
	\caption{Additional examples of navigation instructions in the \coda{} environment. Each instruction is shown with the panoramic view from the starting pose, with the initial heading indicated in red.}
	\label{fig:instr-examples}
\end{figure}

\begin{figure*}[t]
	\begin{center}
		\setlength\tabcolsep{3pt}
		\begin{tabularx}{\linewidth}{ccXX}
			\raisebox{2em}{\rotatebox{90}{\small\textbf{Matterport3D}}} & &
			\includegraphics[width=\linewidth]{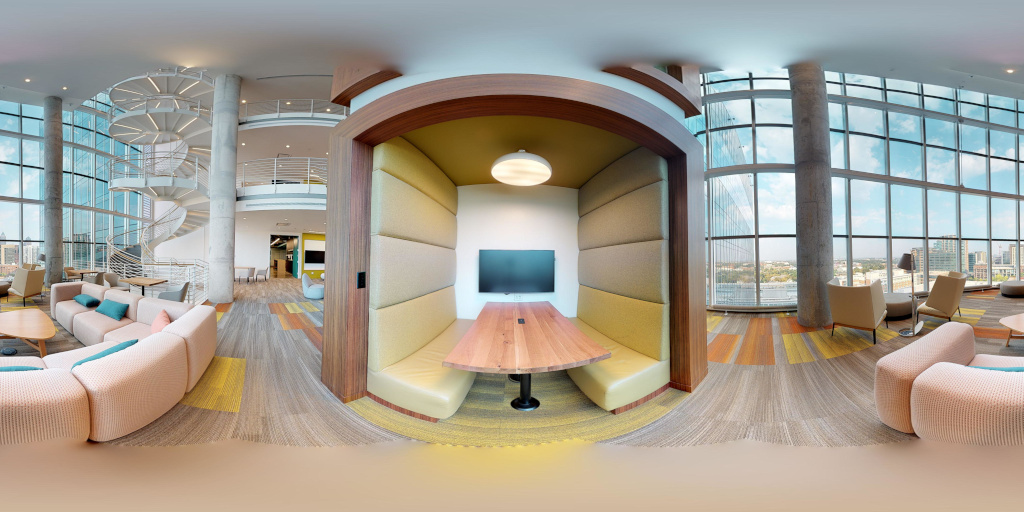} &
			\includegraphics[width=\linewidth]{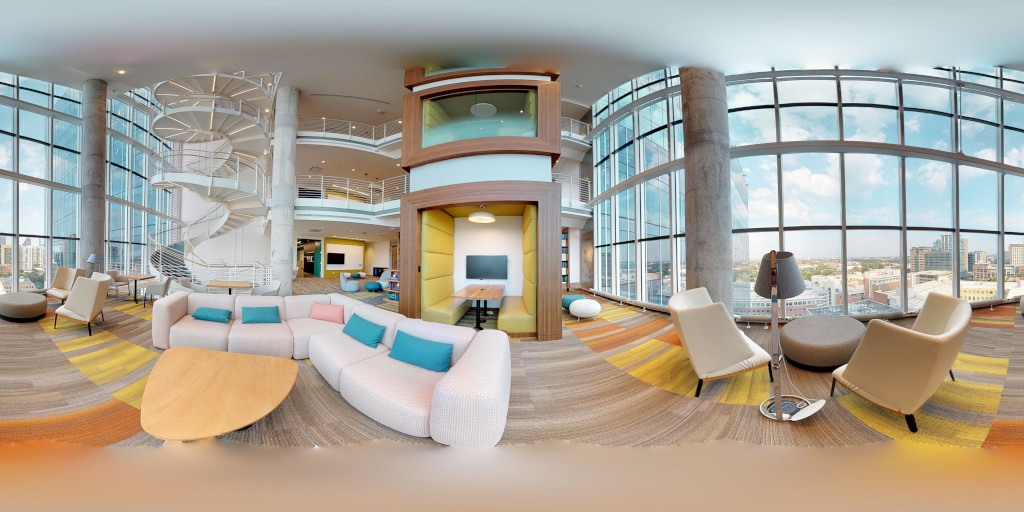} \\
			& \raisebox{2em}{\rotatebox{90}{\parbox{6em}{\centering \scriptsize Day 1}}} &
			\includegraphics[width=\linewidth]{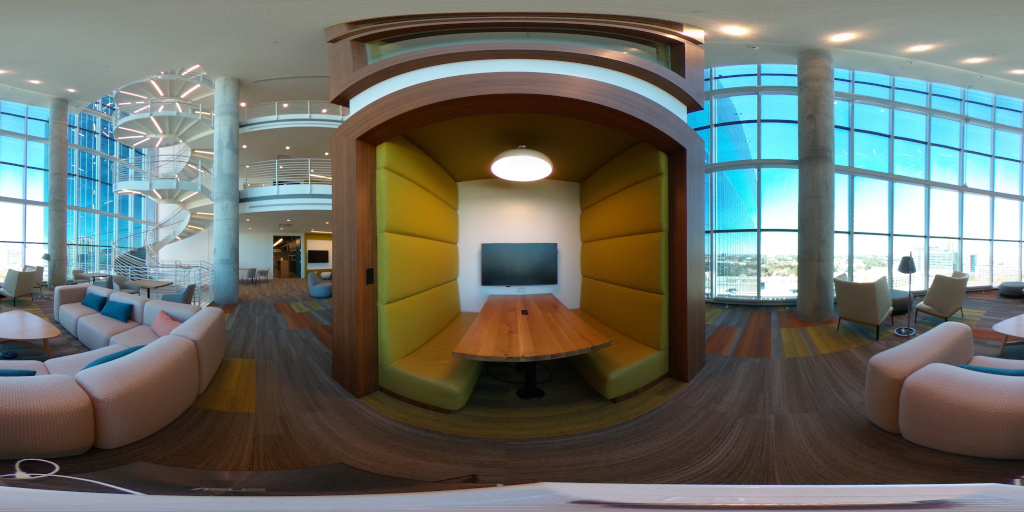} &
			\includegraphics[width=\linewidth]{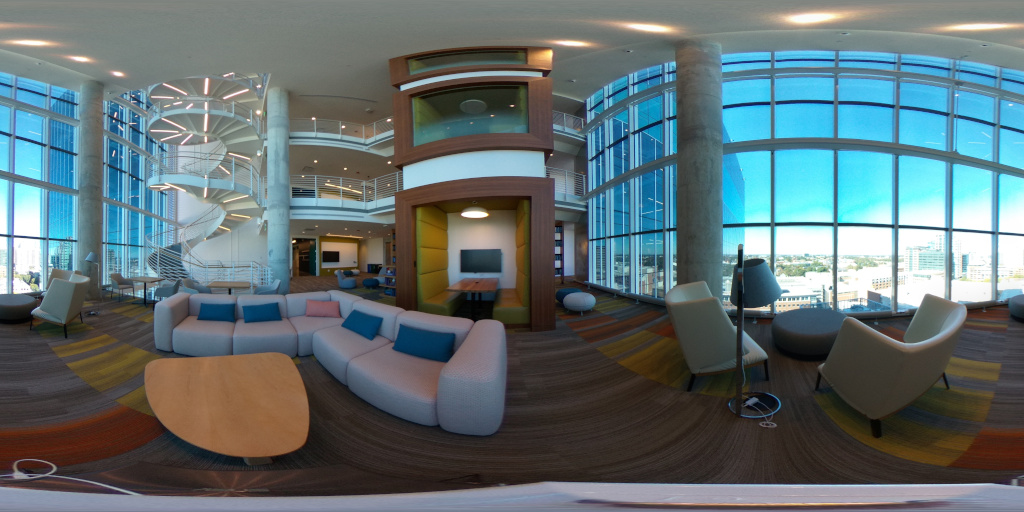} \\
			\raisebox{2em}{\rotatebox{90}{\parbox{6em}{\centering\small\textbf{Ricoh Theta V}}}} &
			\raisebox{2em}{\rotatebox{90}{\parbox{6em}{\centering \scriptsize Day 2}}} &
			\includegraphics[width=\linewidth]{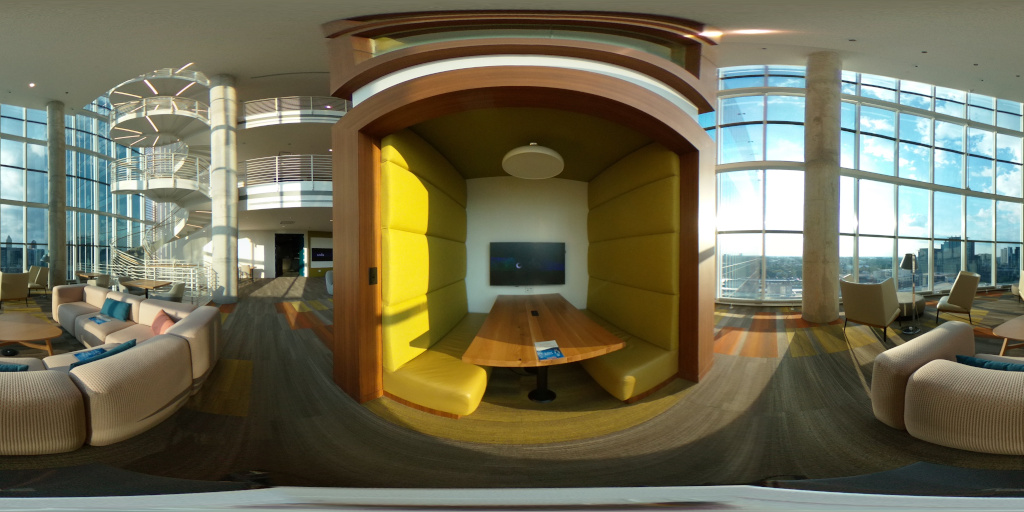} &
			\includegraphics[width=\linewidth]{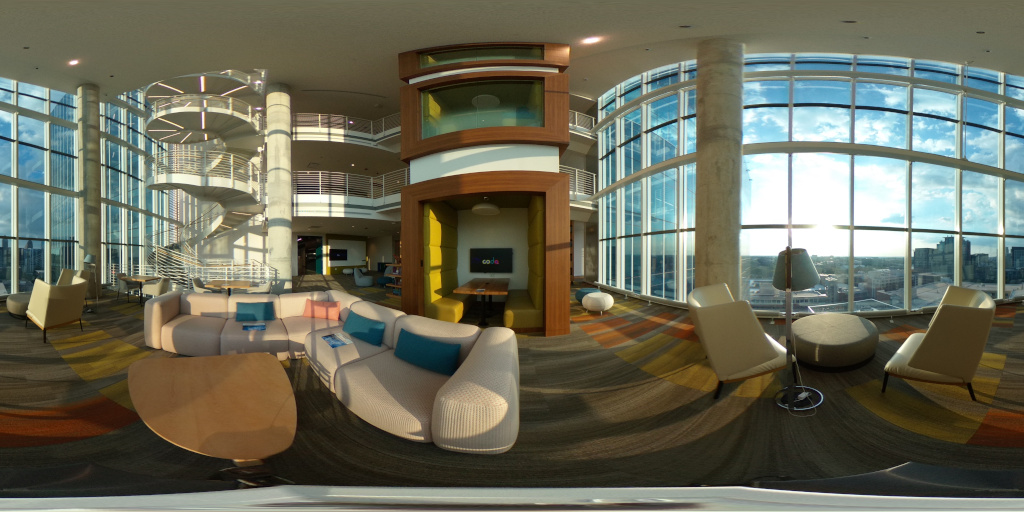} \\
			& \raisebox{2em}{\rotatebox{90}{\parbox{6em}{\centering \scriptsize Day 3}}} &
			\includegraphics[width=\linewidth]{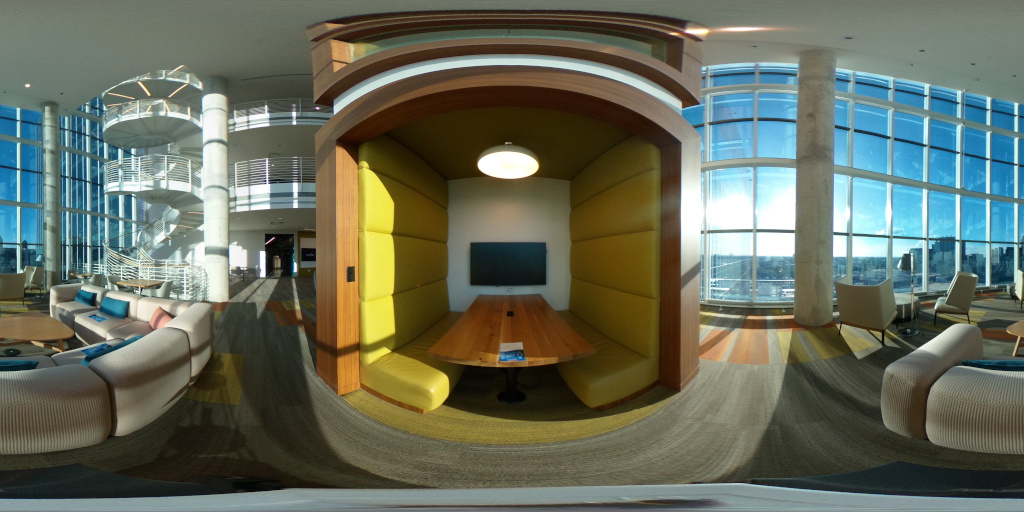} &
			\includegraphics[width=\linewidth]{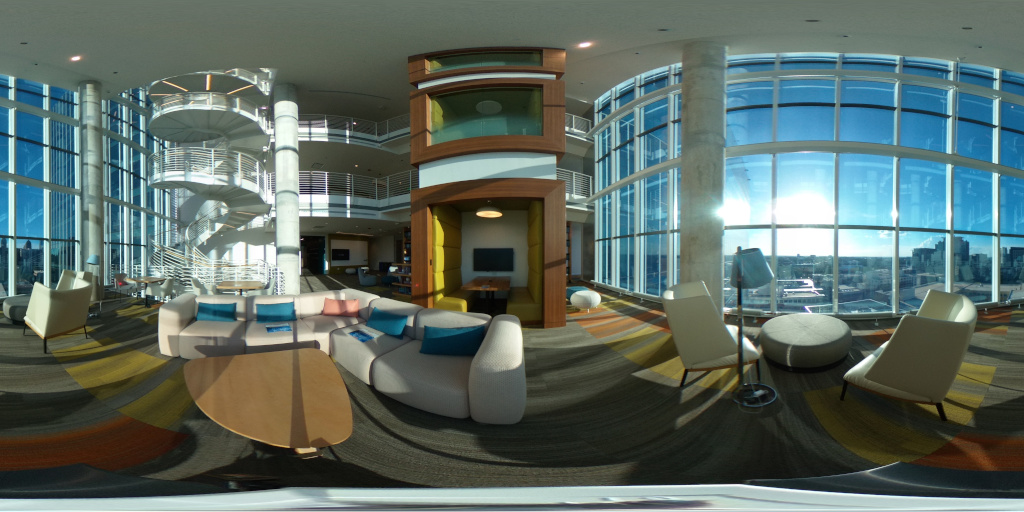} \\
		\end{tabularx}
	\end{center}
	\caption{Panoramic captures in \coda{} from the Matterport3D camera (row 1) and the smaller and cheaper Ricoh Theta V camera (mounted on the robot) collected on three different days (rows 2-4). The robot camera's limited dynamic range and loss of detail as compared to the Matterport3D camera (used for training the VLN agent) is clearly evident. Images collected on different days (with the robot camera) illustrate the variations in shadows, lighting, and precise object placement that confront the robot in the real physical environment.}
	\label{fig:ricoh-supp}
\end{figure*}

\begin{figure}[]
	\begin{center}
		\includegraphics[width=1\linewidth]{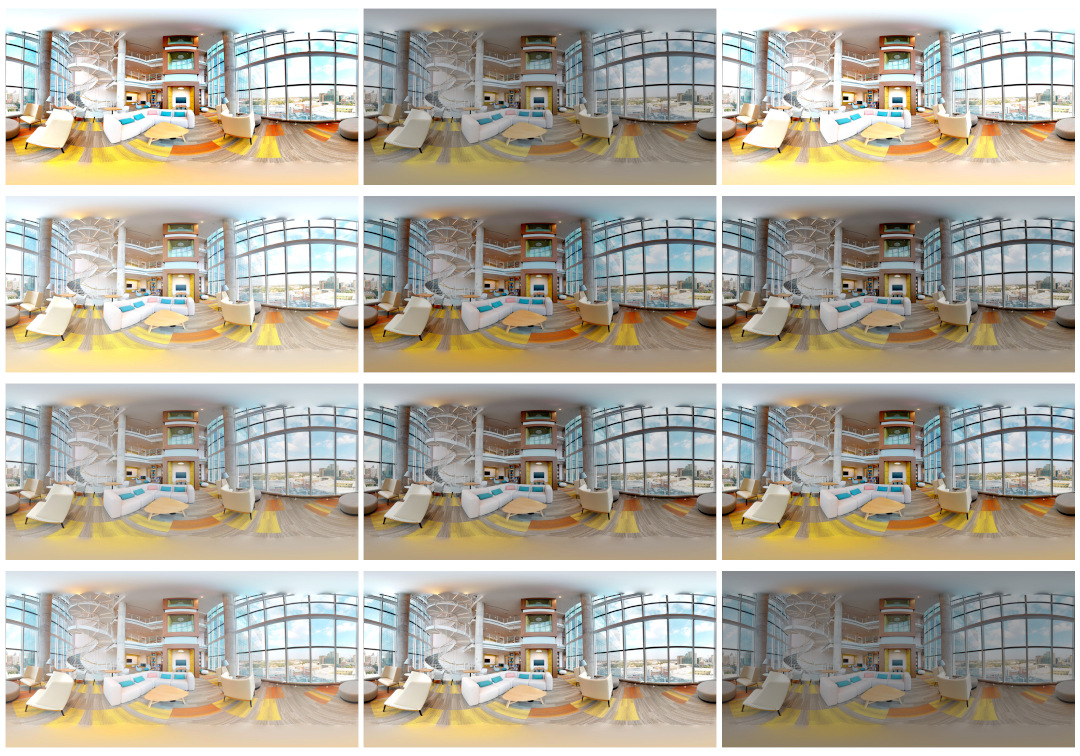}
	\end{center}
	\caption{Examples of the random color jitter applied to each panorama while training the VLN agent to visually adapt to different lighting conditions.}
	\label{fig:jitter}
\end{figure}

\begin{figure}[]
	\begin{center}
		\includegraphics[width=1\linewidth]{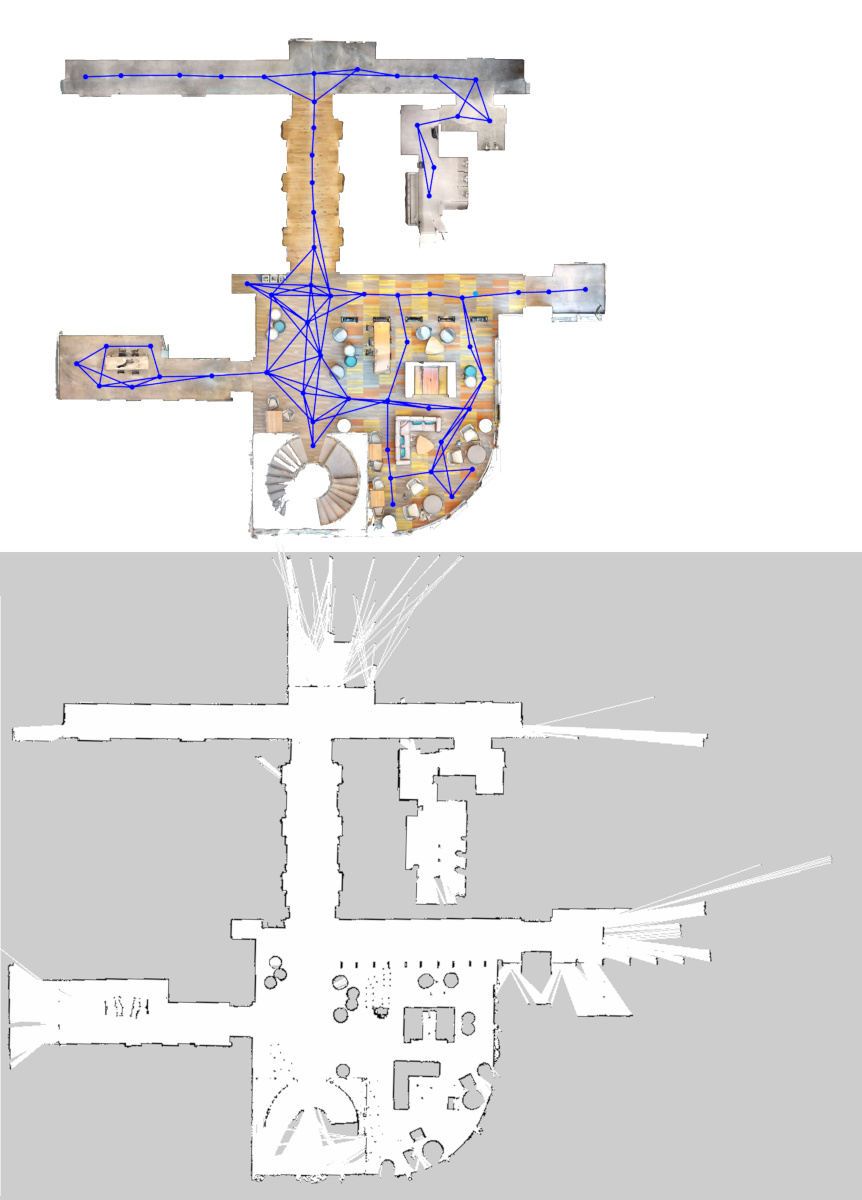}
	\end{center}
	\caption{Floorplan view of \coda{}, showing the Matterport reconstruction and simulator navigation graph (top), and its close alignment to the 2D laser scan used for robot pose tracking (bottom). }
	\label{fig:scan}
\end{figure}

\begin{figure}[t]
	\begin{center}
		\adjustbox{trim={.04\width} {.05\height} {.02\width} {.1\height},clip}%
		{\includegraphics[width=0.52\columnwidth]{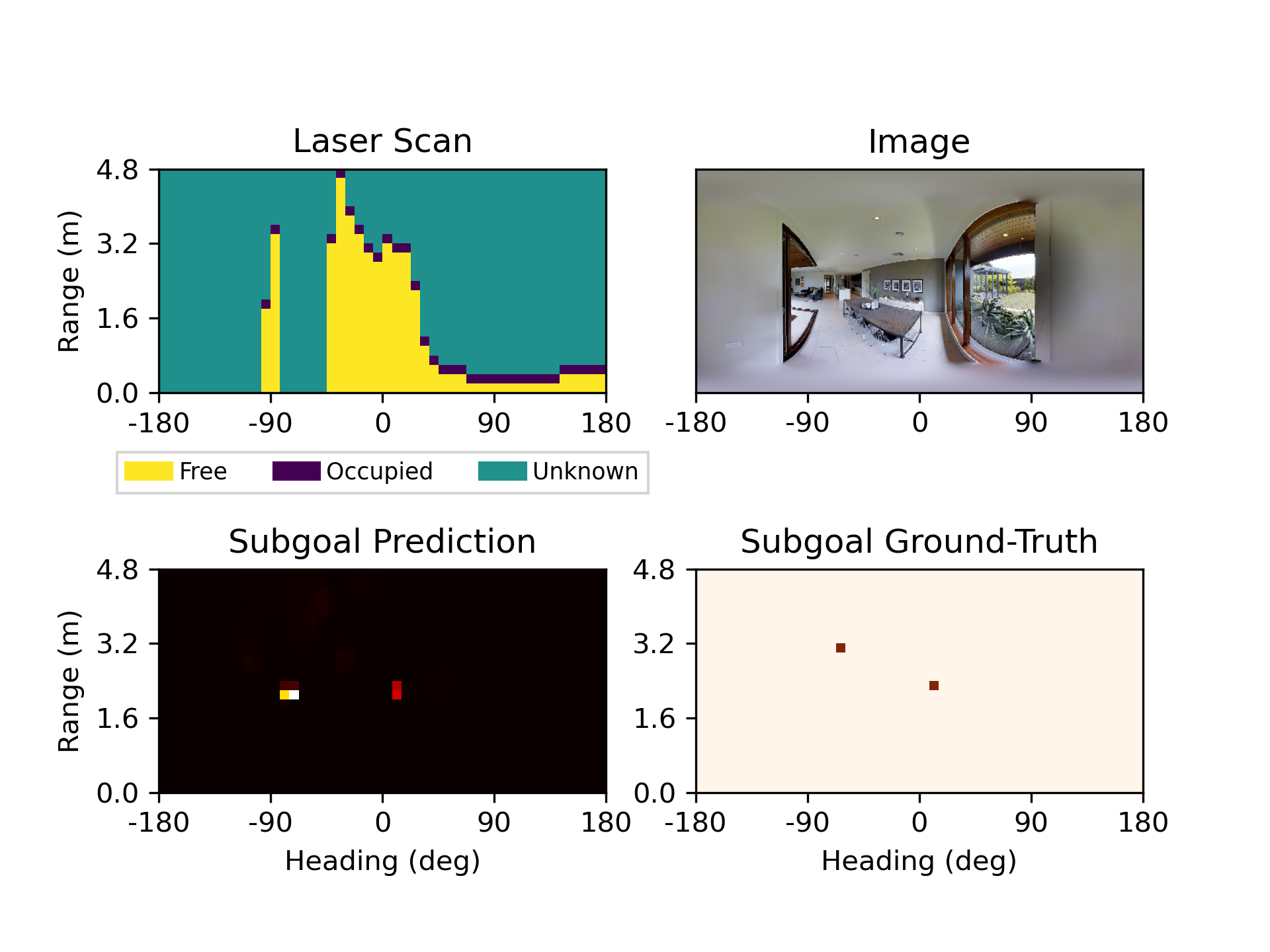}}
		\adjustbox{trim={.01\width} {.05\height} {.06\width} {.1\height},clip}%
		{\includegraphics[width=0.52\columnwidth]{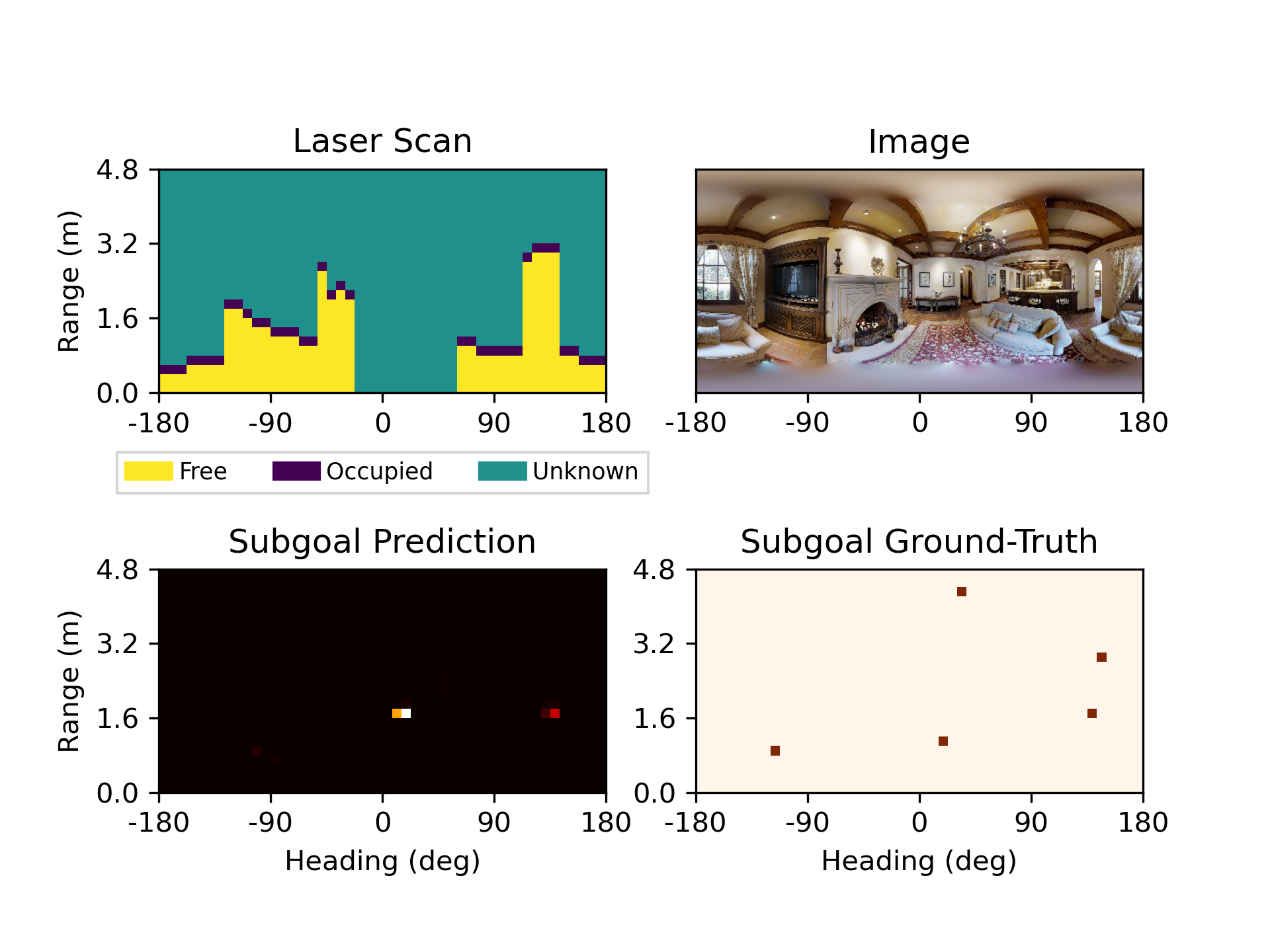}}\\
		\adjustbox{trim={.04\width} {.05\height} {.02\width} {0\height},clip}%
		{\includegraphics[width=0.52\columnwidth]{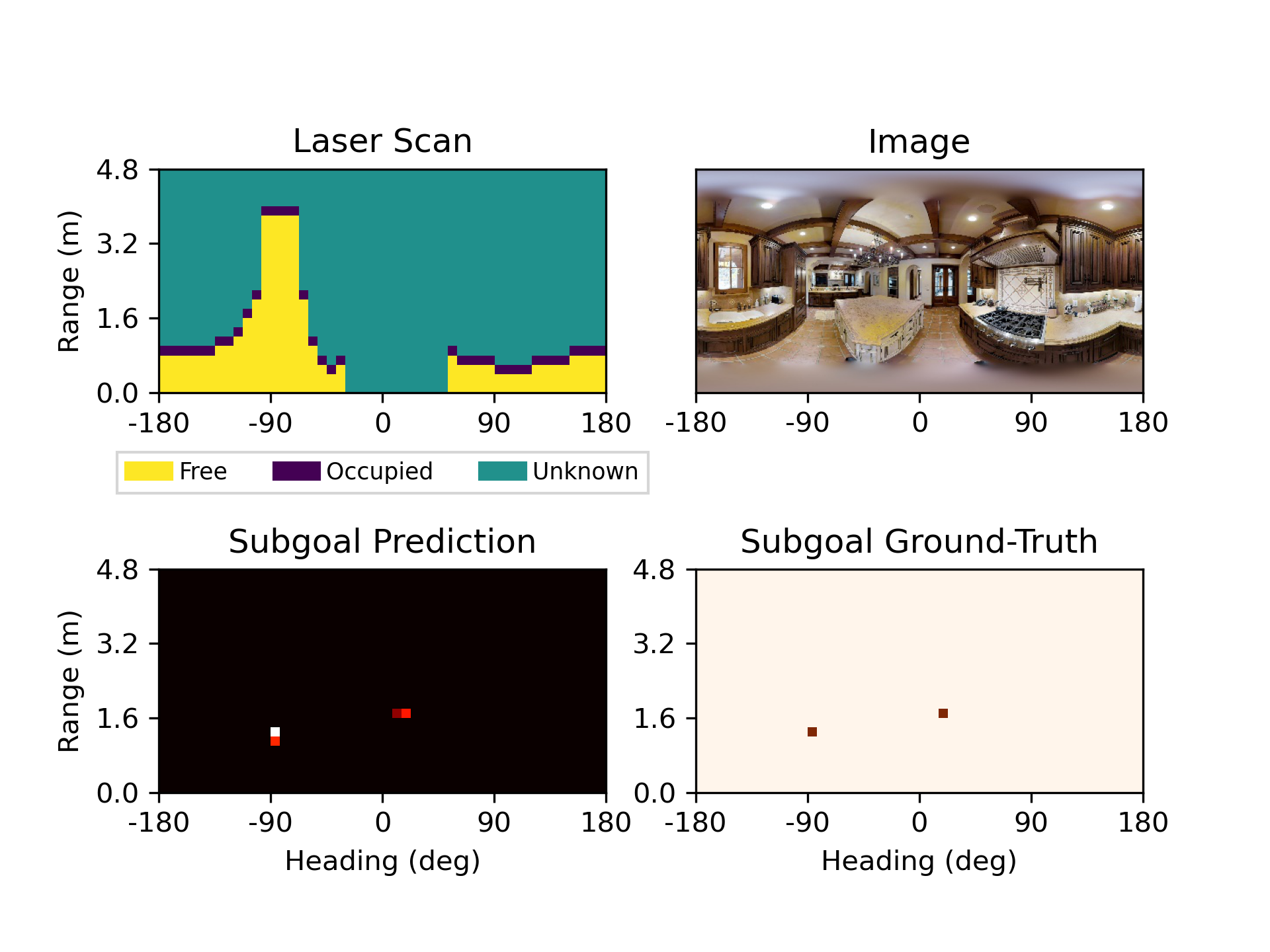}}
		\adjustbox{trim={.01\width} {.05\height} {.06\width} {0\height},clip}%
		{\includegraphics[width=0.52\columnwidth]{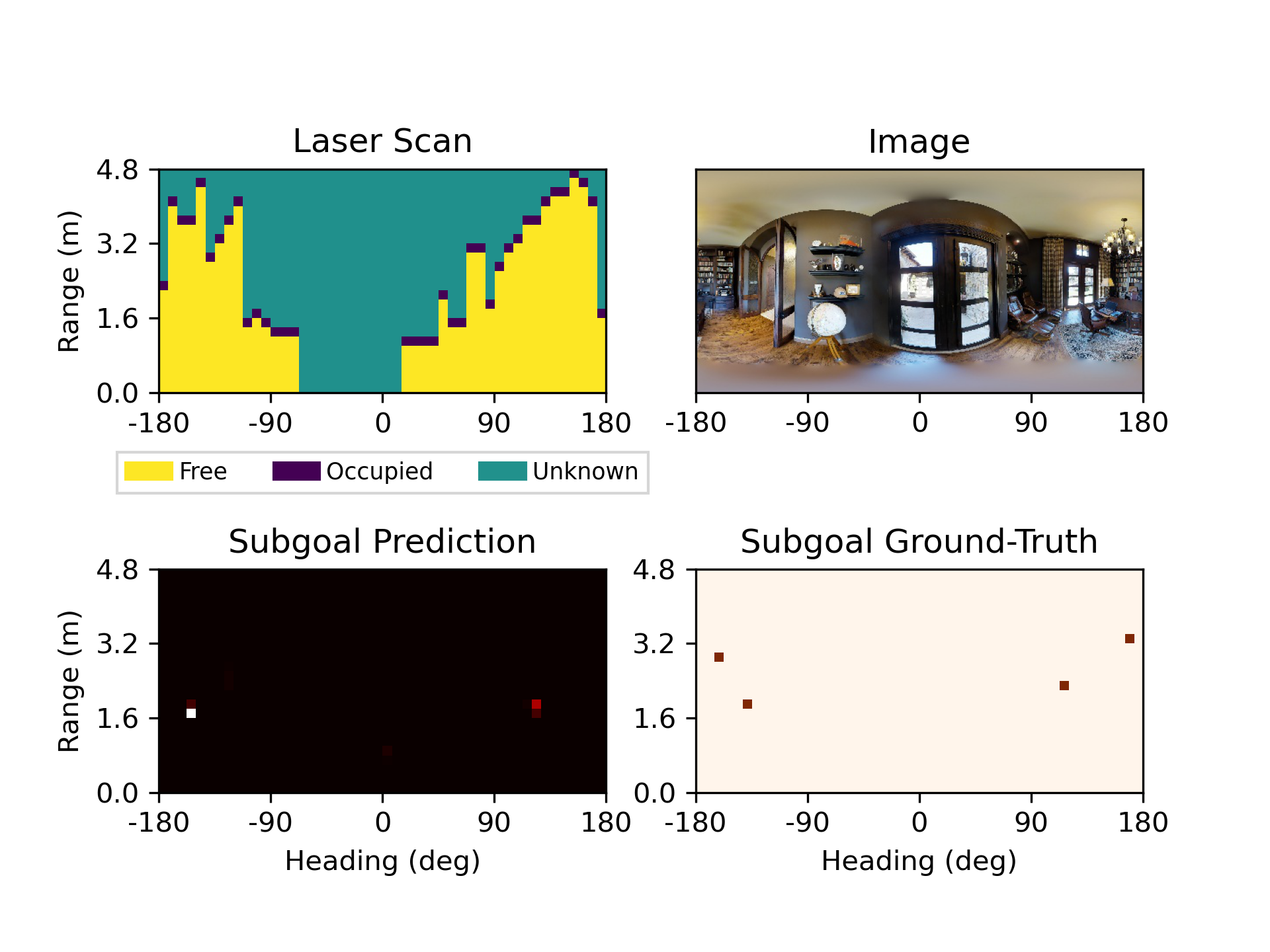}}\\
		\adjustbox{trim={.04\width} {.05\height} {.02\width} {0\height},clip}%
		{\includegraphics[width=0.52\columnwidth]{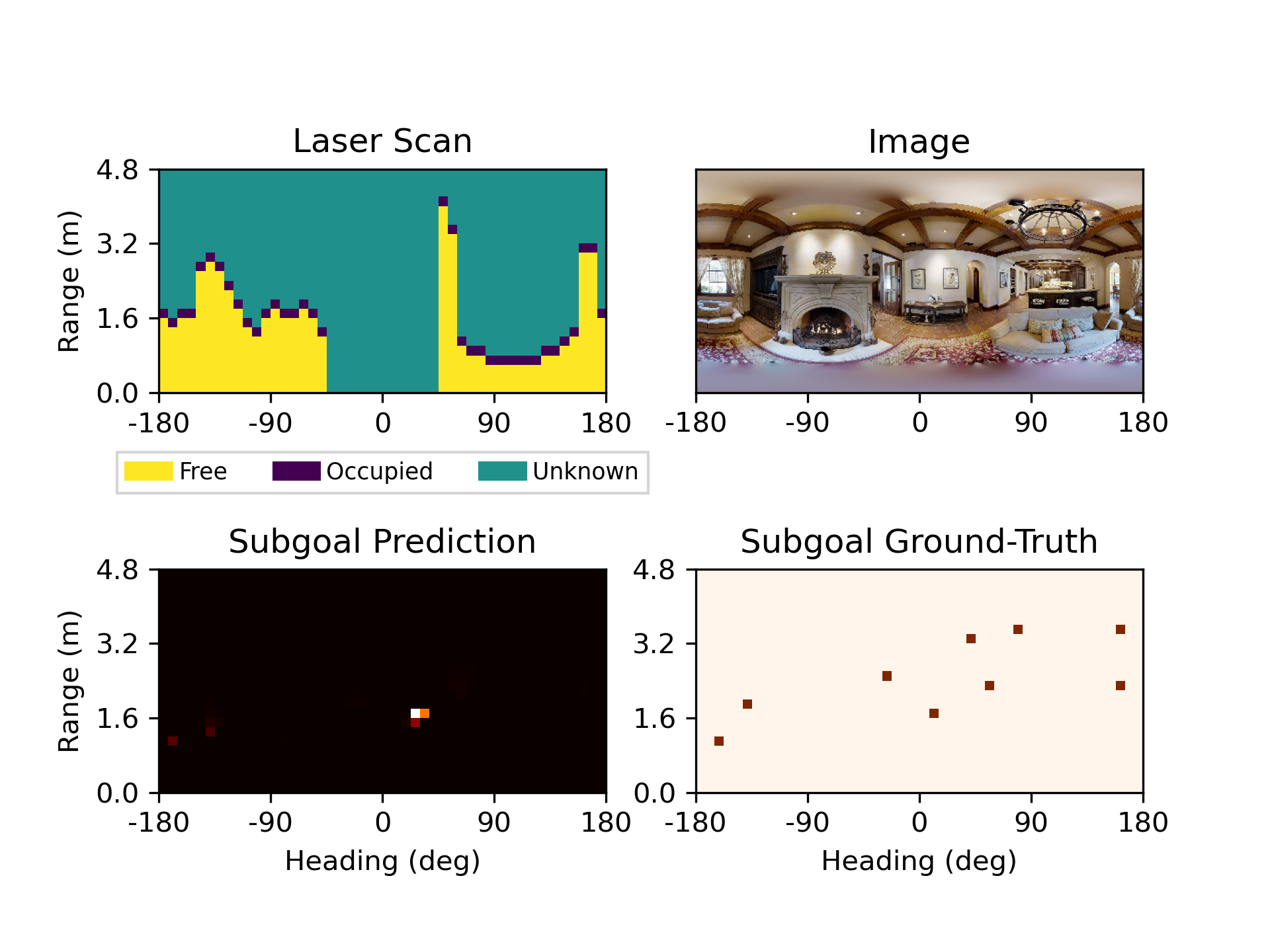}}
		\adjustbox{trim={.01\width} {.05\height} {.06\width} {0\height},clip}%
		{\includegraphics[width=0.52\columnwidth]{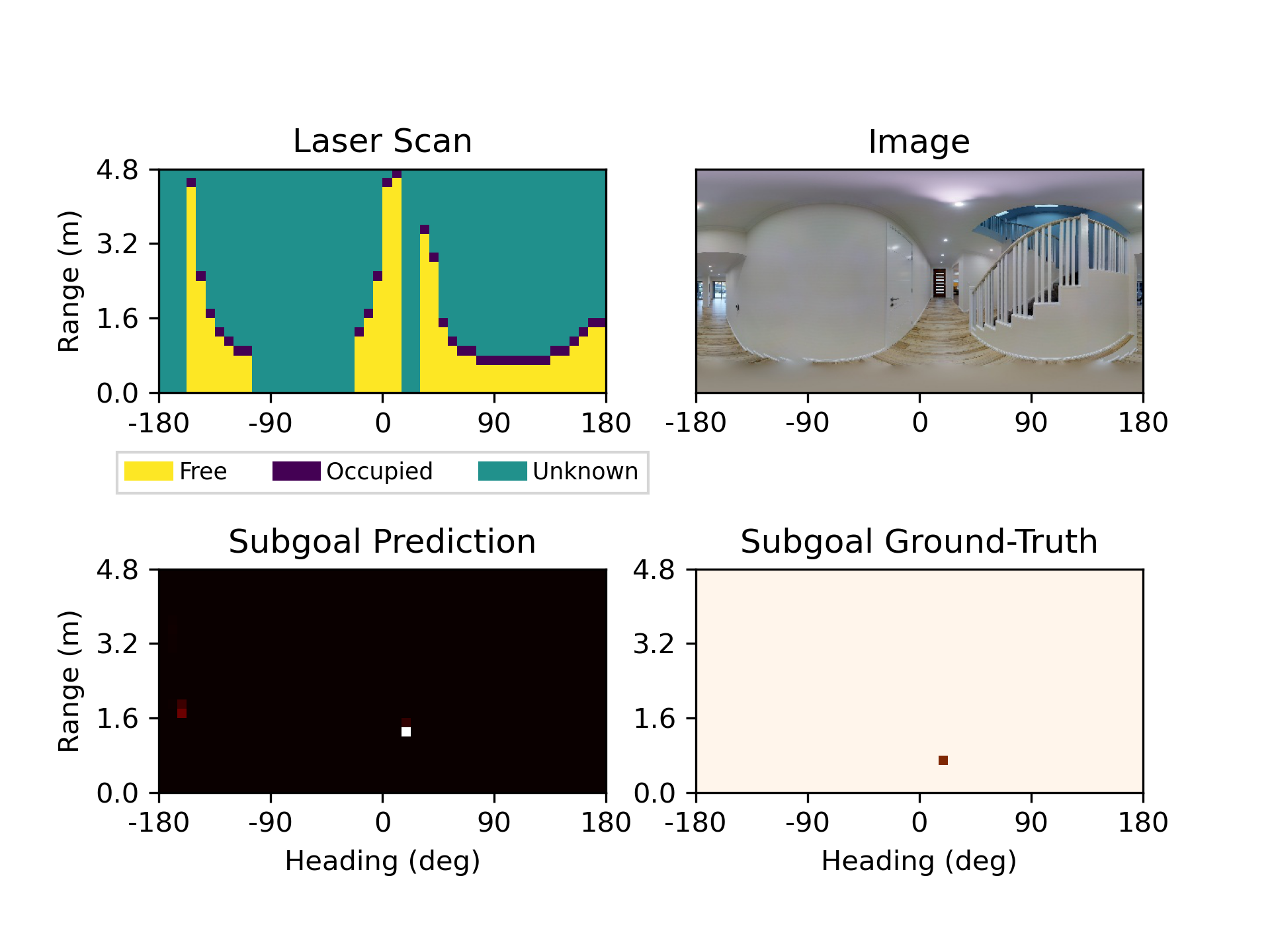}}\\
		\adjustbox{trim={.04\width} {.05\height} {.02\width} {0\height},clip}%
		{\includegraphics[width=0.52\columnwidth]{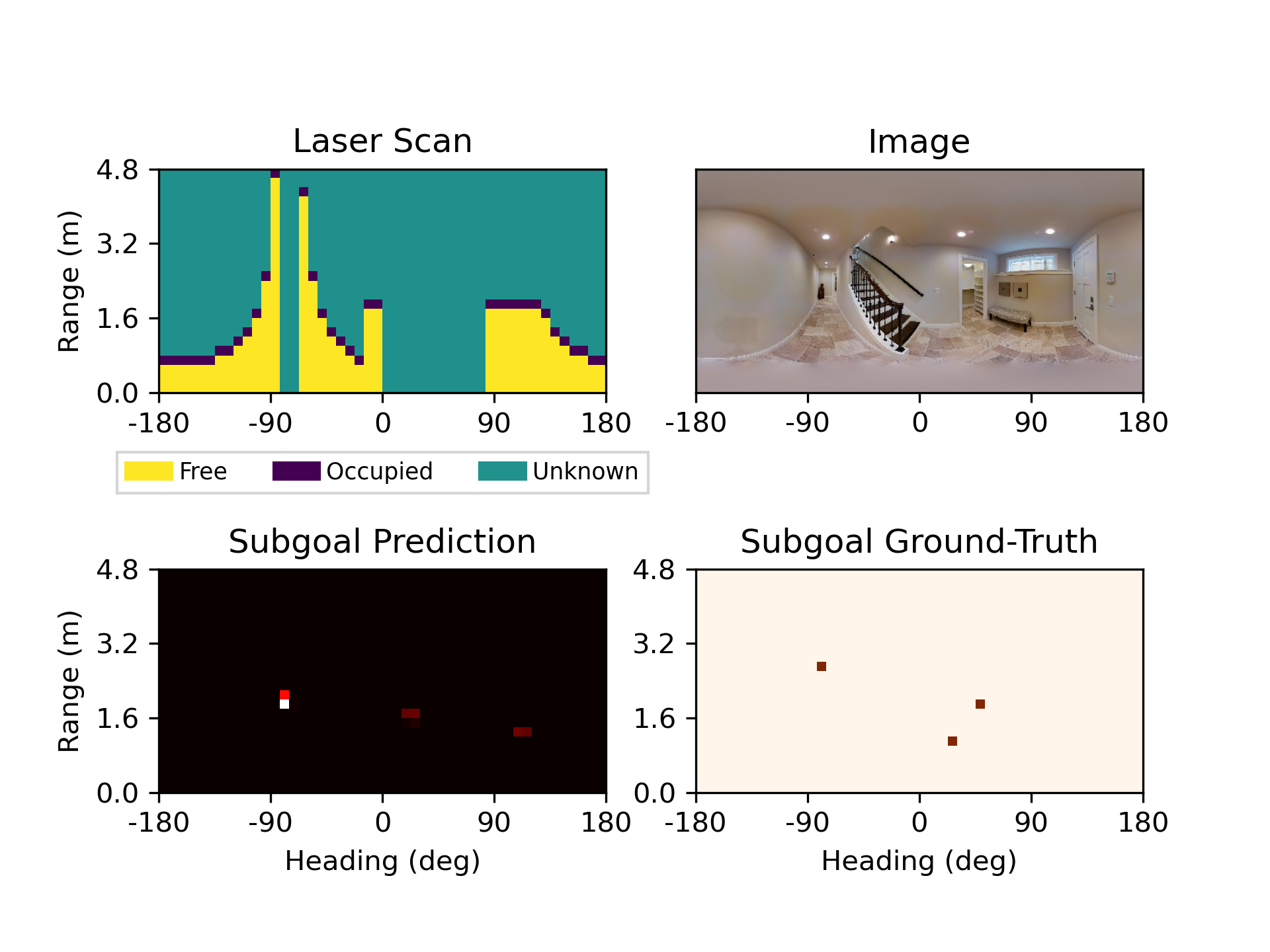}}
		\adjustbox{trim={.01\width} {.05\height} {.06\width} {0\height},clip}%
		{\includegraphics[width=0.52\columnwidth]{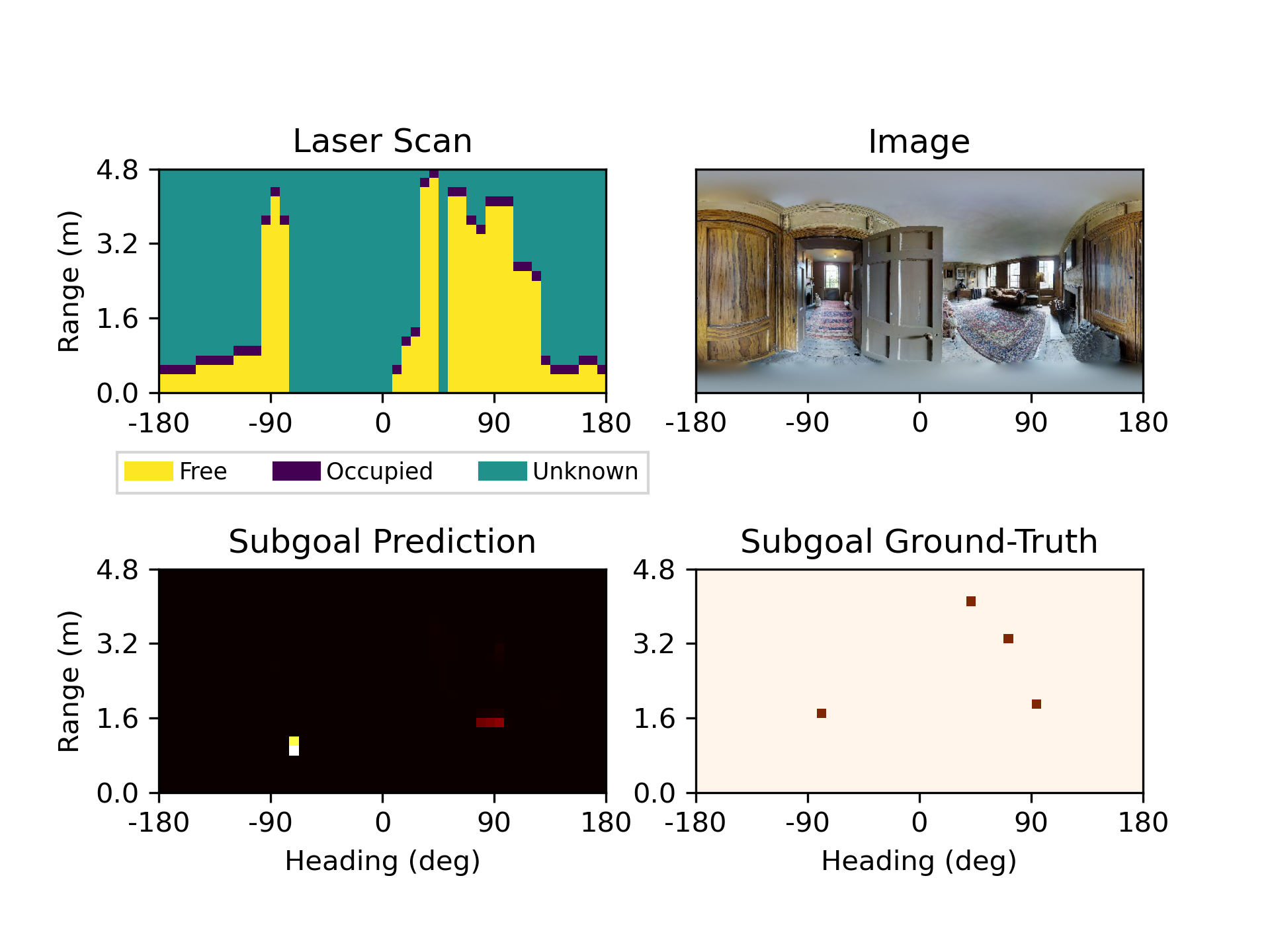}}\\
	\end{center}
	\caption{Waypoint predictions from the subgoal model on 8 randomly selected viewpoints from the Matterport validation set.}
	\label{fig:action-pred-random}
\end{figure}

\begin{figure*}[t]
	\begin{center}
		\small
		\setlength\tabcolsep{10pt}
		\begin{tabularx}{\linewidth}{XX}
			\includegraphics[trim=1cm 0cm 2cm 0, clip,width=1\linewidth]{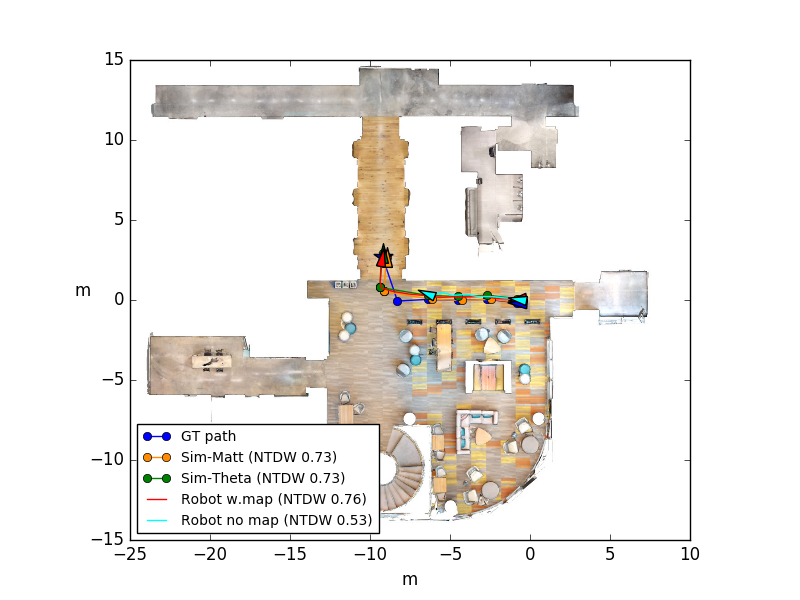} &
			\includegraphics[trim=1cm 0cm 2cm 0, clip,width=1\linewidth]{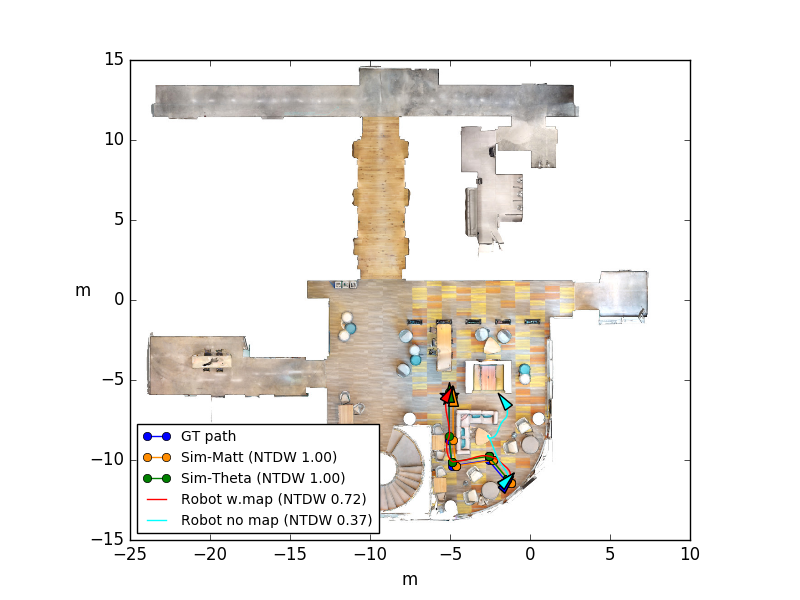} \\
			Instruction: Walk down the hall with the brown shelving units on your left. Turn right into the hallway with the elevators and then stop. 	&
			Instruction: Go between the table bookcase and the sectional sofa on this floor. \\

			\includegraphics[trim=1cm 0cm 2cm 0, clip,width=1\linewidth]{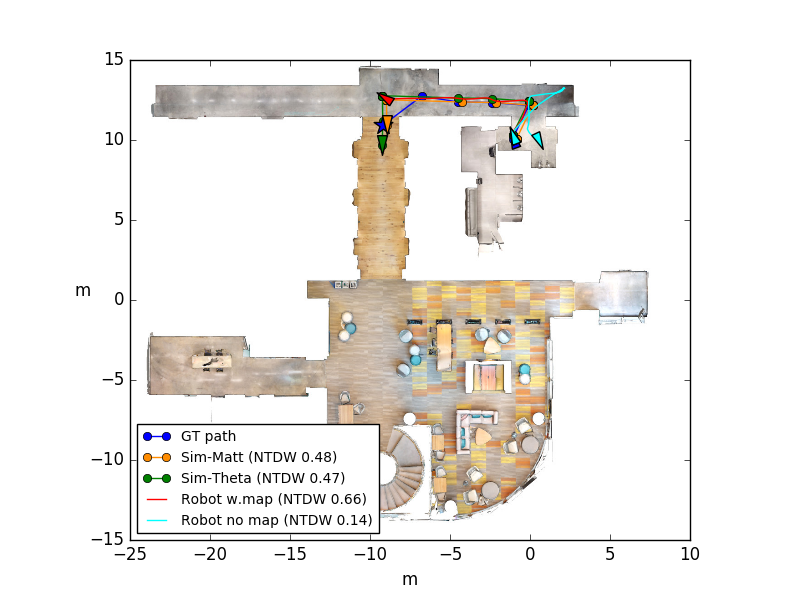} &
			\includegraphics[trim=1cm 0cm 2cm 0, clip,width=1\linewidth]{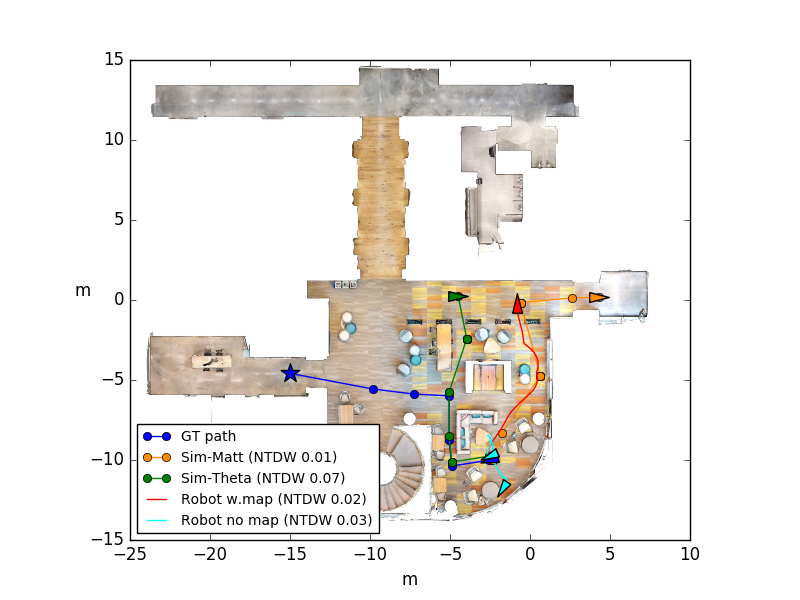} \\
			Instruction: Walk into hallway. Make a left at closed brown door. Walk down hall and make a left and stop by open white doors. 	&
			Instruction: Walk around the back of the large couch. Turn left towards the open green doorway. Walk through the green doors and stop. \\
		\end{tabularx}
	\end{center}
	\caption{Examples of \coda{} trajectories in sim and real for various instructions. While the robot's trajectory often resembles the simulator (top-left), subgoal prediction errors can lead to divergences between the `with map' and `no map' settings (top-right), particularly in areas of the building with floor-to-ceiling glass walls that are not easily detected (bottom-left). In the last example (bottom-right) the agent fails in both sim and real, highlighting the challenging nature of the VLN task. }
	\label{fig:robot-examples}
\end{figure*}

\begin{figure}[]
	\begin{center}
		\includegraphics[trim=1.5cm 0 2cm 0, clip,width=0.7\columnwidth]{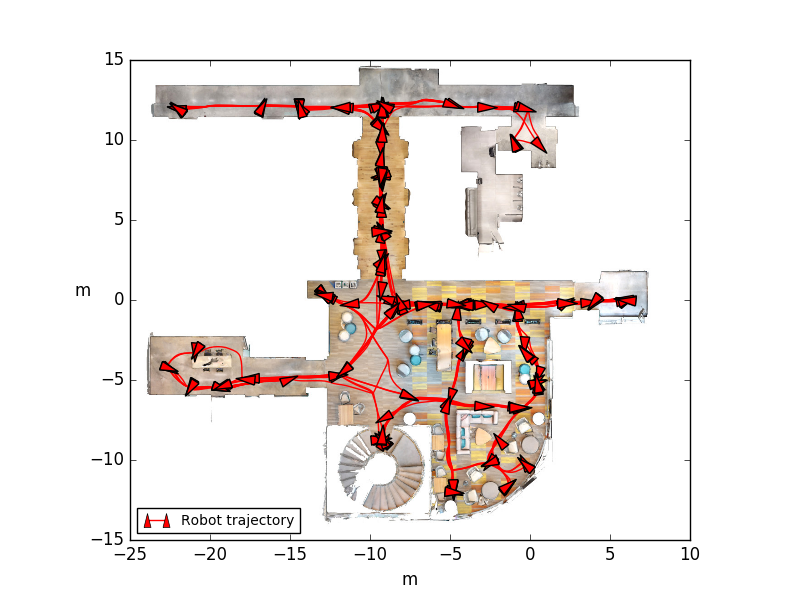}
		\includegraphics[trim=1.5cm 0 2cm 0, clip,width=0.7\columnwidth]{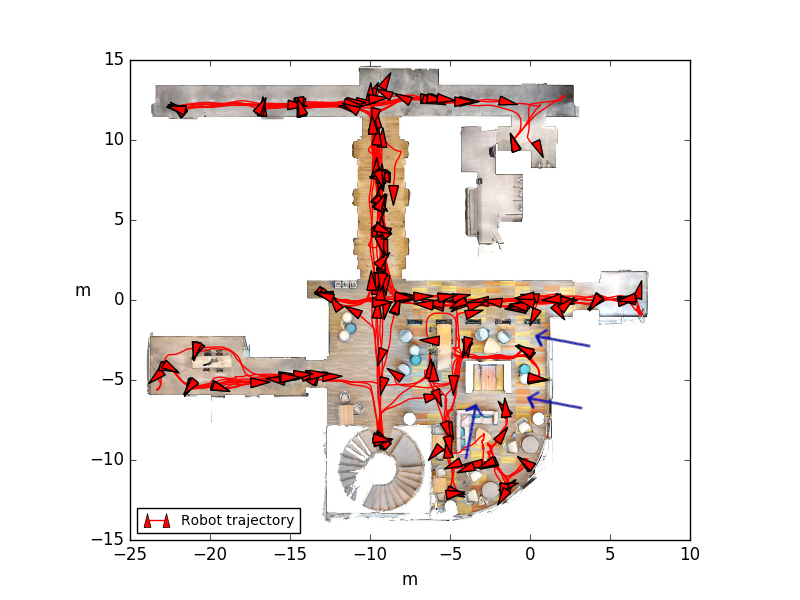}
	\end{center}
	\caption{Illustration of all 111 of trajectories traversed by the robot under the `with map' setting (top) and the `no map' setting (bottom). With a map, the robot traversed the entire space without any collisions or navigation failures. Without a map, certain trajectory segments (highlighted) with blue arrows are never traversed, indicating that the subgoal model failed to predict these waypoints. }
	\label{fig:robot-trials}
\end{figure}

\begin{figure}[]
	\begin{center}
		\small
		\setlength\tabcolsep{10pt}
		\def\arraystretch{2.0}
		\begin{tabularx}{\linewidth}{>{\centering\arraybackslash}X >{\centering\arraybackslash}X}
			\includegraphics[width=\linewidth]{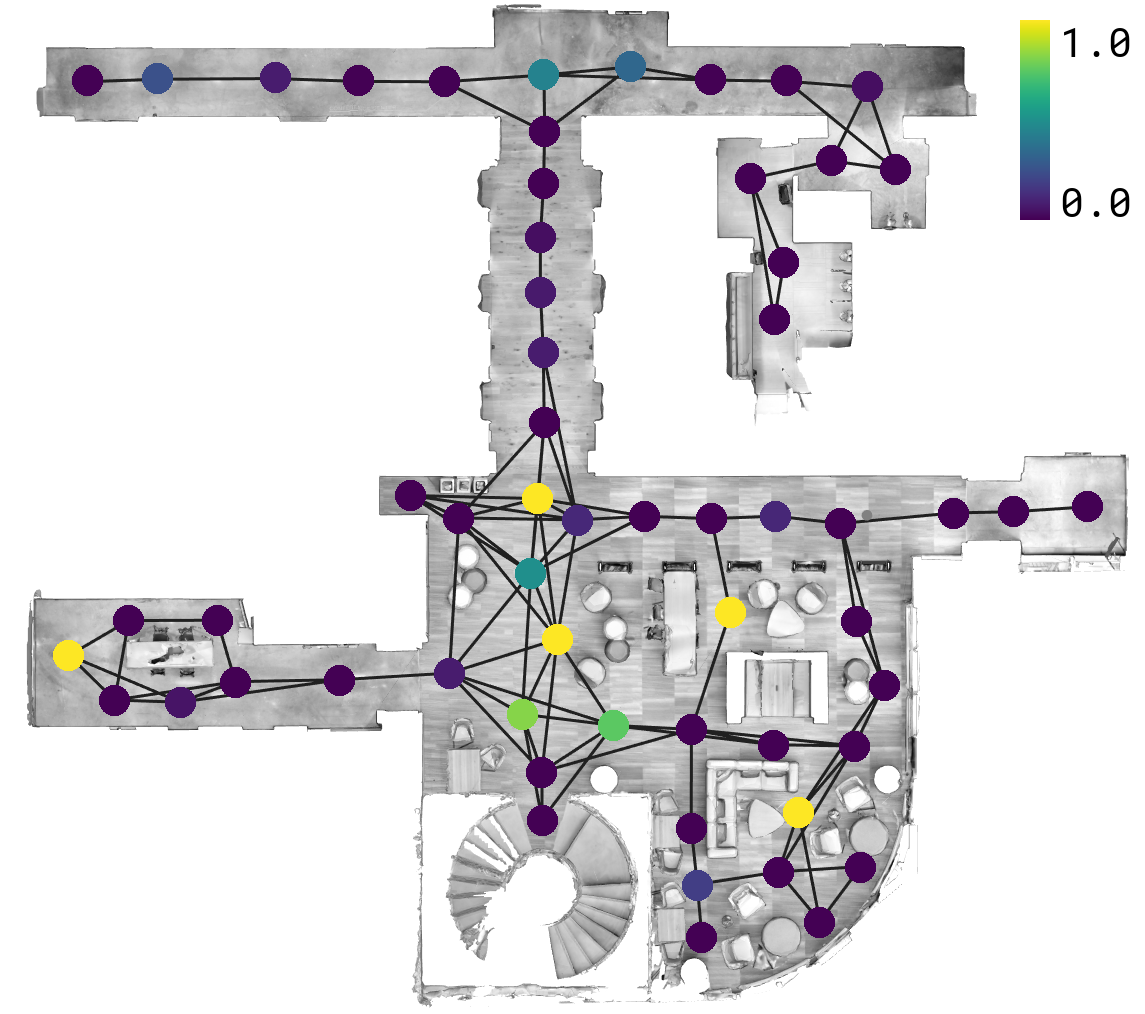} &
			\includegraphics[width=\linewidth]{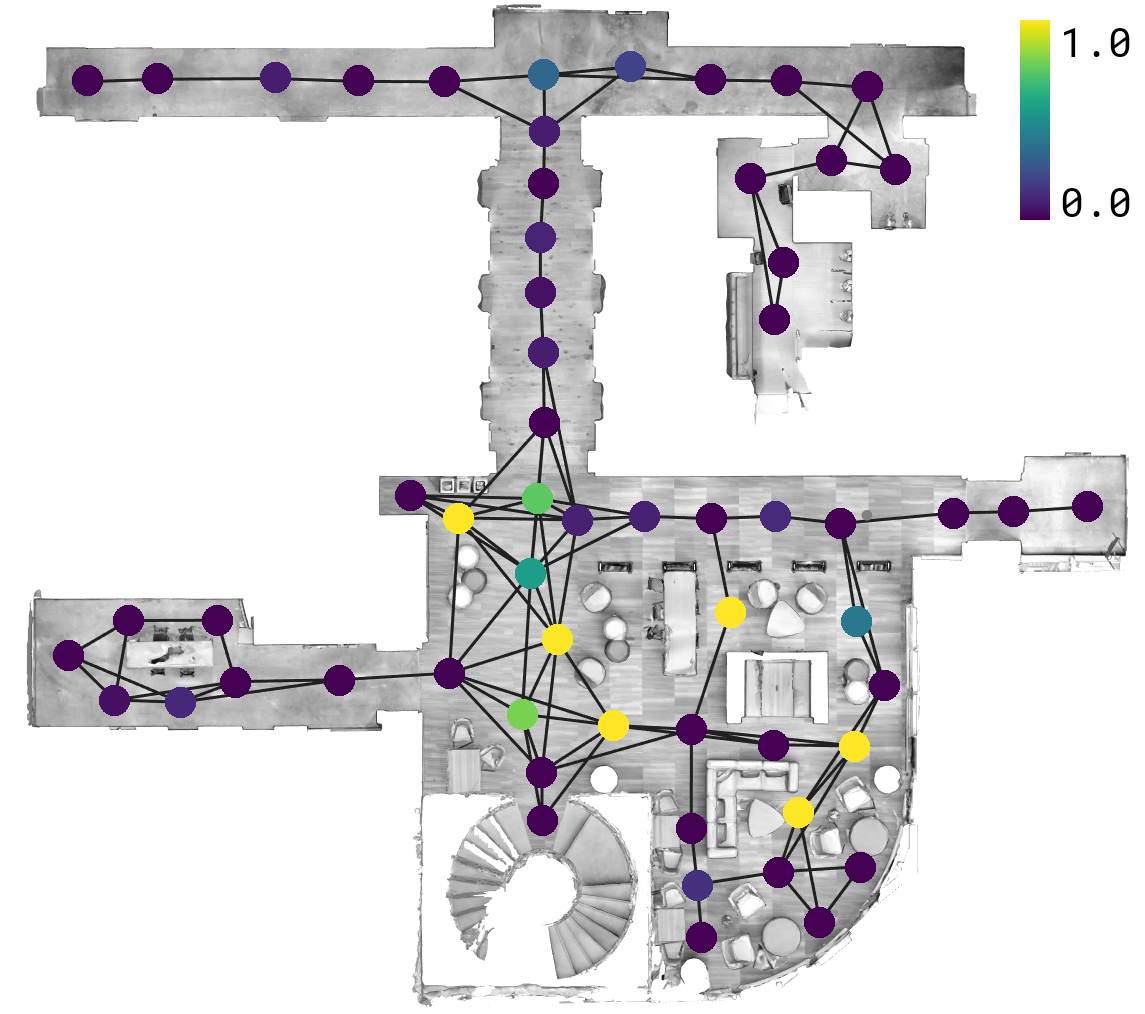} \\
			(a) Matterport3D & (b) Ricoh Theta V Day 1 \\
			\includegraphics[width=\linewidth]{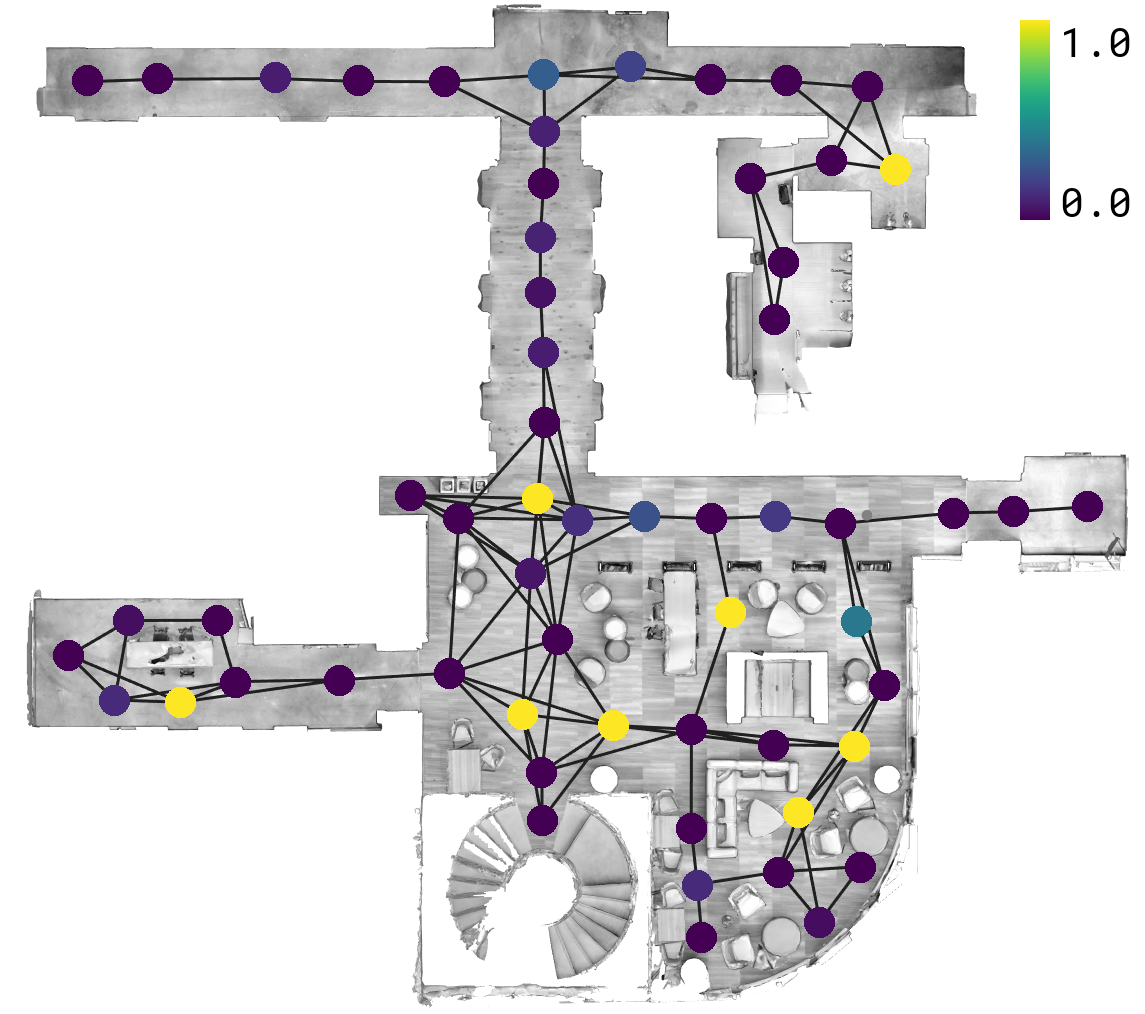} &
			\includegraphics[width=\linewidth]{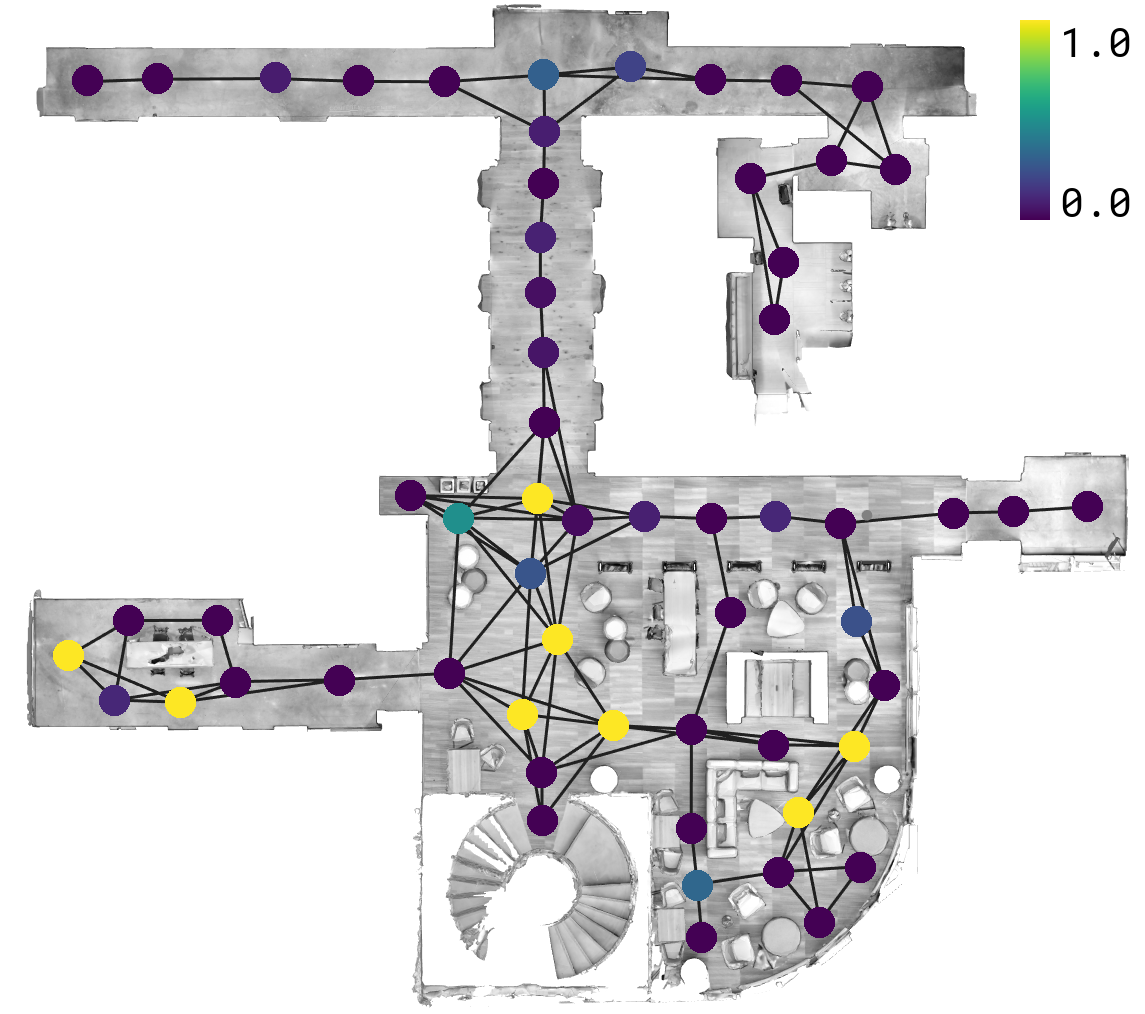} \\
			(c) Ricoh Theta V Day 2 & (d) Ricoh Theta V Day 3 \\
		\end{tabularx}
	\end{center}
	\caption{Illustration of the VLN agent's failure rates (in simulation) at each node in the navigation graph. The failure rates are consistently higher (yellow) in the bottom right of the map across all four data collections with the Matterport3d and Ricoh Theta V cameras.}
\end{figure}

\begin{figure}[]
	\begin{center}
		\small
		\def\arraystretch{2.0}
		\begin{tabularx}{\linewidth}{>{\centering\arraybackslash}X}
			\includegraphics[width=0.4\columnwidth]{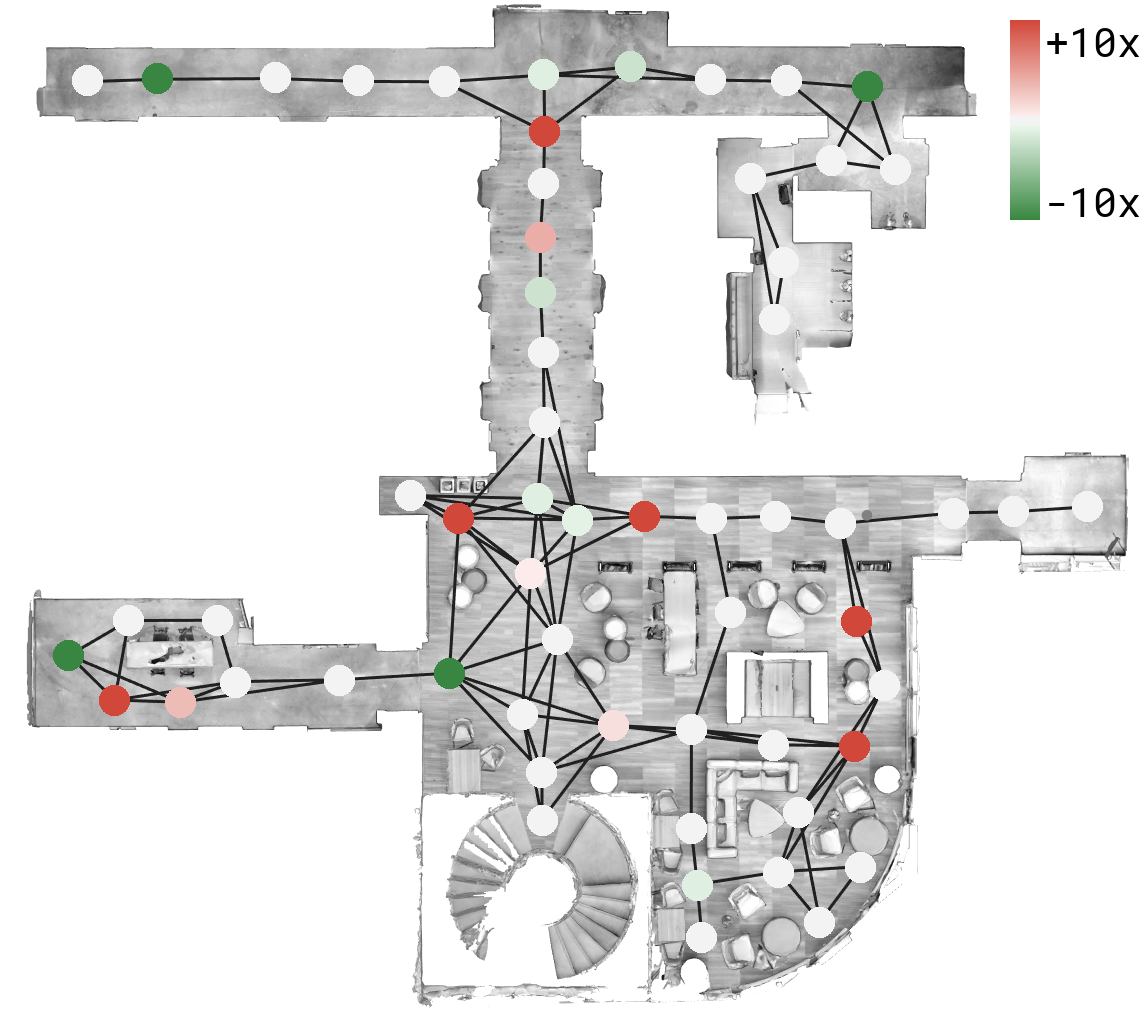} \\
			(a) Failure Rate Ratio - Ricoh Theta V Day 1 to Matterport3D \\
			\includegraphics[width=0.4\columnwidth]{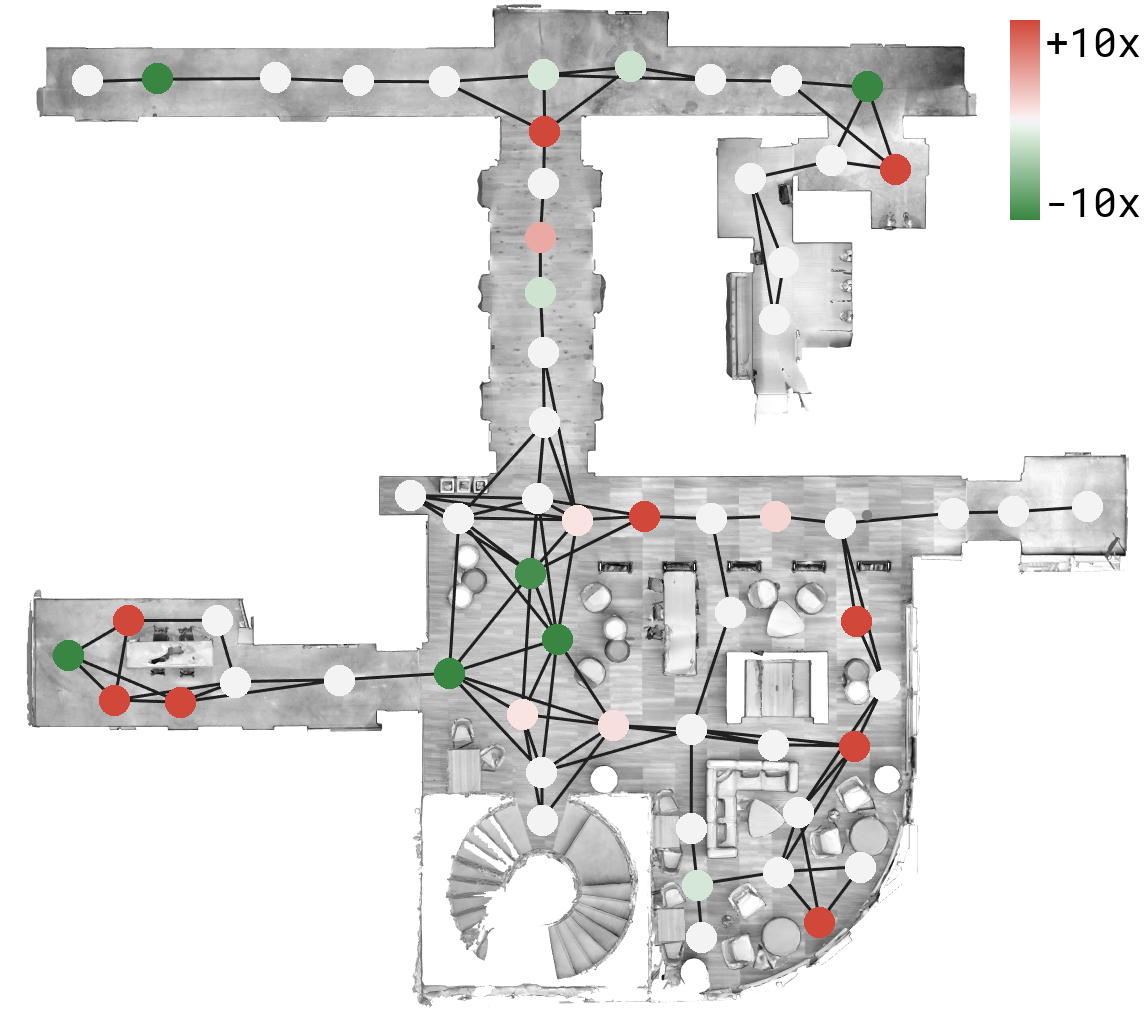} \\
			(b) Failure Rate Ratio - Ricoh Theta V Day 2 to Matterport3D \\
			\includegraphics[width=0.4\columnwidth]{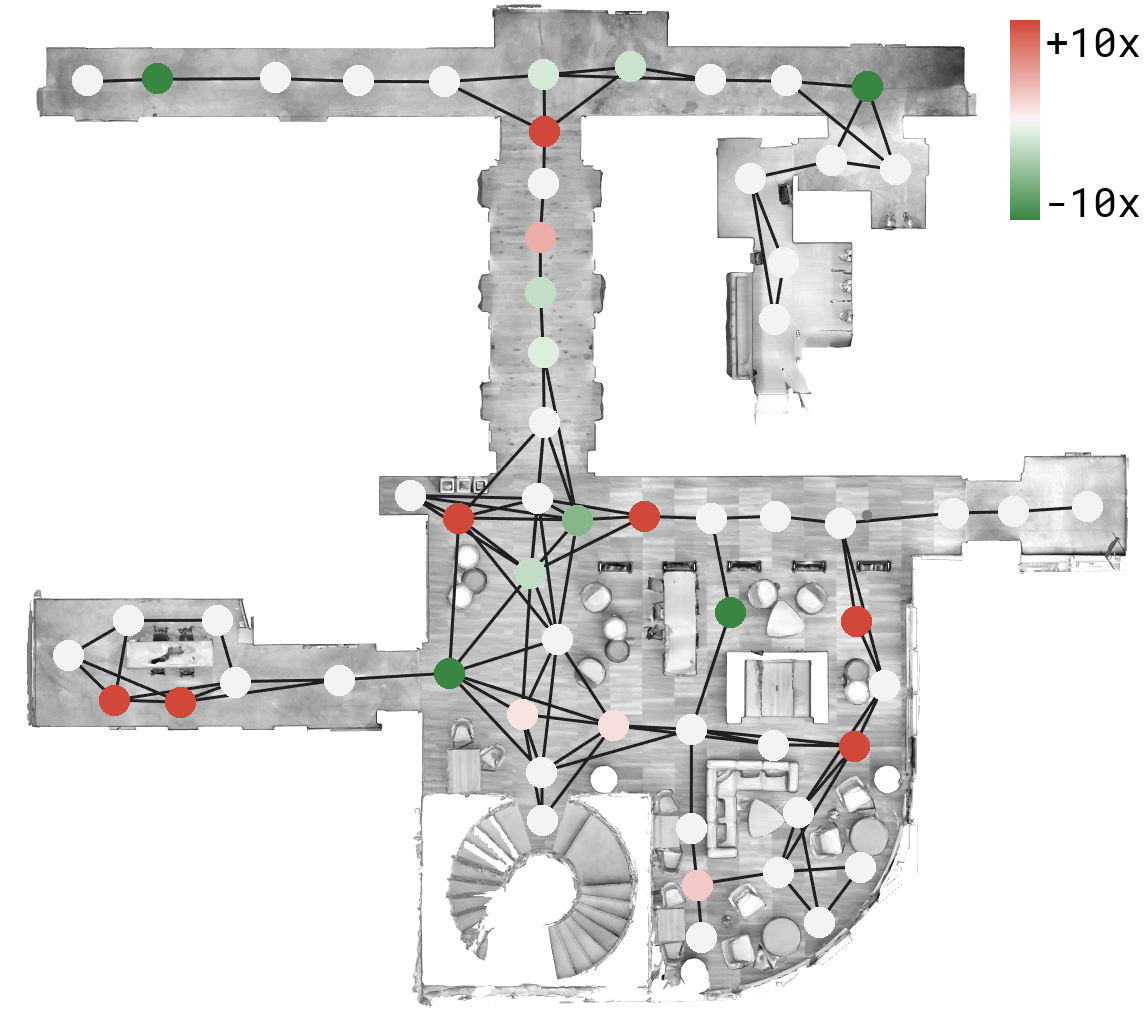} \\
			(c) Failure Rate Ratio - Ricoh Theta V Day 3 to Matterport3D \\
		\end{tabularx}
	\end{center}
	\caption{Illustration of the $\log$ of the ratio between the failure rates with the Ricoh Theta V camera and the Matterport3D camera. A positive ratio, illustrated in red, indicates that the VLN agent was more likely to fail when processing data from the Ricoh Theta V camera. A negative ratio (in green) indicates the opposite. Across all three days, the agent was more likely to fail at nodes to the top and bottom of the elevator area and at nodes near the glass windows in the open space in the bottom right. Surprisingly, there are also some nodes (in green) at which the agent consistently performs better using the Ricoh Theta V panoramas.}
\end{figure}

\end{document}